\documentclass[letterpaper, 10 pt, journal, twoside]{IEEEtran}

\IEEEoverridecommandlockouts                              % This command is only needed if
\usepackage{hyperref}
\usepackage{multirow}
\usepackage{romannum}
\usepackage{graphicx}
\usepackage{amsmath}
\usepackage{amssymb}
\usepackage{breqn}
\usepackage{subcaption}
\usepackage{algorithm}
\usepackage{algorithmicx,algpseudocode}
\algrenewcommand\algorithmicindent{0.6em}
\usepackage{pgfplots}
\usepackage{booktabs, dcolumn}

\usepackage{soul}
\soulregister\cite7
\soulregister\ref7

\pgfplotsset{
every axis/.append style={
  axis line style={->}, % arrows on the axis
  xlabel near ticks,
  ylabel near ticks,
  legend style={font=\scriptsize},
  label style={font=\scriptsize},
  tick label style={font=\scriptsize},
  title style={font=\scriptsize}
  }
}
\pgfplotsset{compat=newest}
\pgfplotsset{plot coordinates/math parser=false}
\newlength\figureheight
\newlength\figurewidth

\makeatletter
\let\NAT@parse\undefined
\makeatother
\usepackage[numbers,sort,compress]{natbib}
\usepackage{hyperref}
\usepackage[font=footnotesize]{caption}
\usepackage{xcolor,colortbl}
\begin{document}

\title{
MFuseNet: Robust Depth Estimation with Learned Multiscopic Fusion
}

\author{Weihao Yuan, Rui Fan, Michael Yu Wang, and Qifeng Chen % <-this % stops a space
\thanks{Manuscript received: September, 10, 2019; Revised December, 6, 2019; Accepted January, 24, 2020.}%Use only for final RAL version
\thanks{This paper was recommended for publication by Editor Cadena L. Cesar upon evaluation of the Associate Editor and Reviewers' comments.} %Use only for final RAL version
\thanks{Authors are with the Hong Kong University of Science and Technology, Hong Kong SAR, China. W. Yuan ({\tt\footnotesize weihao.yuan@connect.ust.hk}) and R. Fan are with the Department of Electronic and Computer Engineering. M. Y. Wang is with the Department of Mechanical and Aerospace Engineering and the Department of Electronic and Computer Engineering. Q. Chen ({\tt\footnotesize cqf@ust.hk}) is with the Department of Computer Science and Engineering and the Department of Electronic and Computer Engineering.}%
\thanks{Digital Object Identifier (DOI): see top of this page.}
}

\markboth{IEEE Robotics and Automation Letters. Preprint Version. Accepted January, 2020}
{Yuan \MakeLowercase{\textit{et al.}}: MFuseNet: Robust Depth Estimation with Learned Multiscopic Fusion}

\maketitle
% % \thispagestyle{empty}
% % \pagestyle{empty}

%%%%%%%%%%%%%%%%%%%%%%%%%%%%%%%%%%%%%%%%%%%%%%%%%%%%%%%%%%%%%%%%%%%%%%%%%%%%%%%%

\pagenumbering{arabic}

%%%%%%%%%%%%%%%%%%%%%%%%%%%%%%%%%%%%%%%%%%%%%%%%%%%%%%%%%%%%%%%%%%%%%%%%%%%%%%%%
% !TEX root =  ../main.tex
\begin{abstract}

We design a multiscopic vision system that utilizes a low-cost monocular RGB camera to acquire accurate depth estimation. Unlike multi-view stereo with images captured at unconstrained camera poses, the proposed system controls the motion of a camera to capture a sequence of images in horizontally or vertically aligned positions with the same parallax. In this system, we propose a new heuristic method and a robust learning-based method to fuse multiple cost volumes between the reference image and its surrounding images. To obtain training data, we build a synthetic dataset with multiscopic images. The experiments on the real-world Middlebury dataset and real robot demonstration show that our multiscopic vision system outperforms traditional two-frame stereo matching methods in depth estimation. Our code and dataset are available at \url{https://sites.google.com/view/multiscopic}.

\end{abstract}

\begin{IEEEkeywords}
Visual learning, deep learning in robotics and automation, computer vision for automation, depth estimation, multiscopic vision.
\end{IEEEkeywords}
% !TEX root =  ../main.tex

\section{INTRODUCTION}

% application
%Reconstructing a 3 dimensional scene is an important perception problem which is the premise for many robotic applications, such as manipulation, exploration and navigation. There are many sensors that can be used to do this like RGB camera, depth sensor and LiDAR, of which the RGB camera is of the lowest cost and can obtain dense depth image. But the reconstruction from 2-D images to 3-D scene is always one of the most active research areas and not solved.
\IEEEPARstart{U}{nderstanding} surrounding 3-dimensional (3D) environments is an essential perception task for numerous robotic applications, including manipulation, exploration, and navigation \cite{biswas2012depth, ye2019tightly, yuan2019reinforcement, yuan2019end}. Robots usually rely on accurate depth estimation of a scene to avoid obstacles and manipulate objects. In industrial environments, a color camera is usually installed on moving agents such as autonomous ground vehicles (AGV) and robot arms that can control the camera movement. Therefore, can we obtain highly accurate depth maps with a monocular camera by controlling the camera motion?

For depth estimation, we typically utilize depth sensors such as stereo cameras, structured-light sensors, and time-of-flight sensors, but these depth sensors are usually expensive compared to a single RGB camera. Researchers have been working on depth estimation with a single monocular RGB camera, but monocular depth estimation is still far from perfect. We demonstrate that if we can control the motion of an RGB camera and capture images at well-controlled positions and orientations, monocular depth estimation can be significantly improved.

% motivation
%Most methods using RGB cameras are built upon the principles of stereo matching. 
The principle of stereo matching is one of the fundamental techniques for depth estimation with two cameras. A stereo sensor is equipped with two cameras displaced horizontally so that the corresponding pixels in the two cameras are on the same horizontal line. Thus stereo matching can estimate a disparity map that represents the position differences between corresponding pixels in stereo images \cite{scharstein2002taxonomy}. On the other hand, structure from motion (SFM) \cite{koenderink1991affine} and multi-view stereo (MVS) \cite{seitz2006comparison} do not constrain the camera poses so that the pixel correspondence is not on a fixed line, which makes finding pixel correspondences more challenging. 
% While corresponding pixels are restricted on fixed lines, stereo matching has become the key algorithm for stereo cameras and structured-light 3D scanners. 

%We extend the idea of stereo matching to multiscopic vision to obtain high-quality depth with a single camera with regulated motion, in which more constraints can be enforced in reconstructed depth maps. %Both the magnitude and direction of the pixels disparities can be controlled such that we can search the correspondence easily. 
%We strictly control the camera motion horizontally or vertically to capture multiple images so that the correspondence of a pixel is restricted to a fixed line. %This can enable accurate depth estimation in various industrial applications.
%In numerous industrial environments, a color camera is usually installed on moving agents such as autonomous ground vehicles (AGV) and robot arms that can control the camera movement. 

\begin{figure}[]
\centering
  \includegraphics[width=0.8\columnwidth, trim={0cm 0cm 0cm 0cm}, clip]{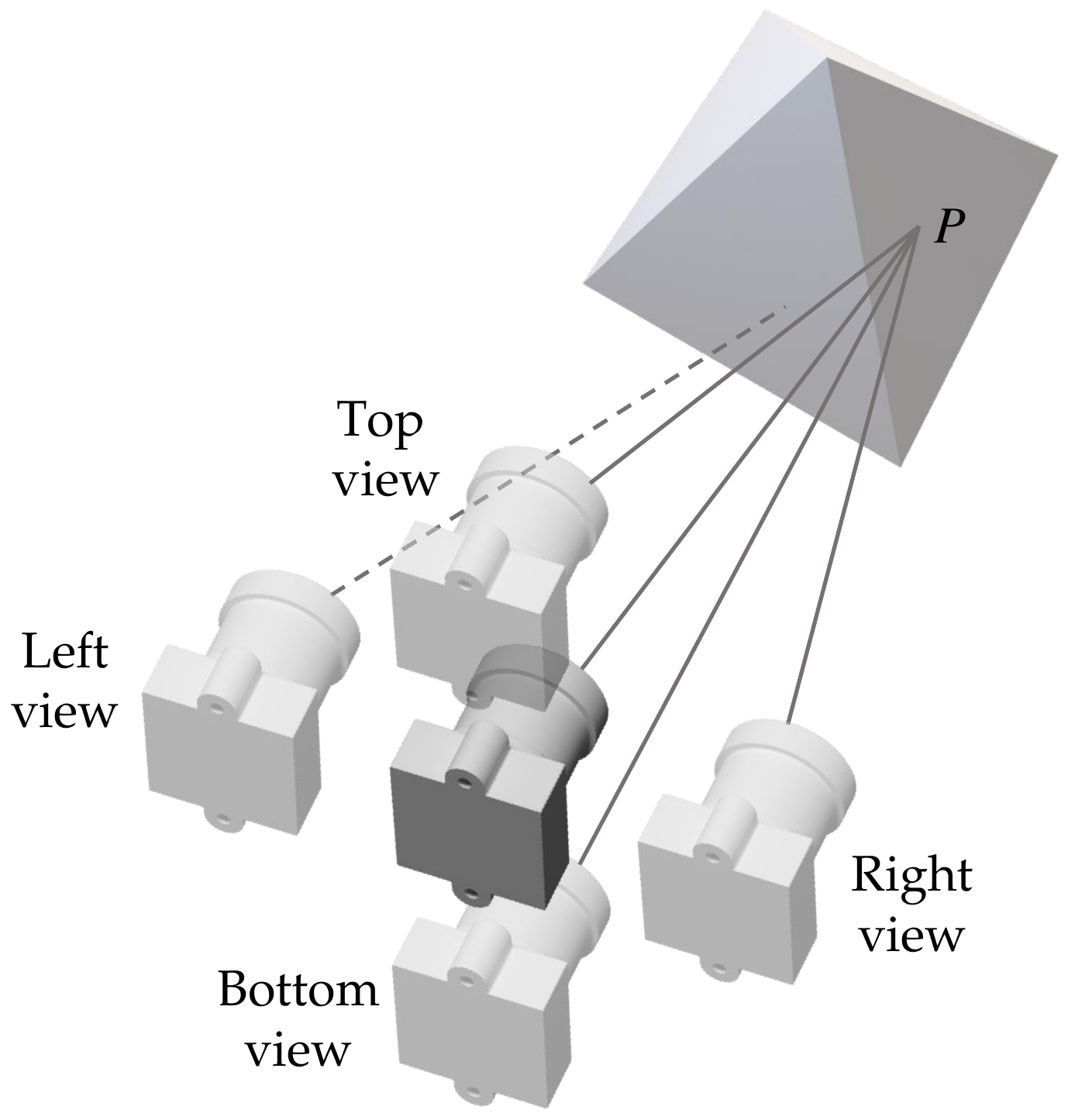}
\caption{Multiscopic vision system. A camera is moved under control so that the captured images are co-planar with the same parallax. The point $P$ on the pyramid can not be perceived from the left view but can be seen from other views. The depth map for the center view can be obtained by fusing the matching costs between the center view and all other views.}
\label{fig:camera}
\vspace{-0.5cm}
\end{figure}

% what this paper is about
We extend the idea of stereo matching to multiscopic vision to obtain high-quality depth with a single camera with regulated motion, in which more constraints can be enforced in reconstructed depth maps. We study depth estimation with a single camera by taking multiple images at specified camera poses. We refer to the problem of depth estimation with multiple images captured at aligned camera locations as multiscopic vision, as an analog to stereo vision with two horizontally aligned images. Inspired by the principle of stereo vision that depth estimation with two perfectly aligned images is relatively easier than with two images with arbitrary unknown camera poses, we believe capturing multiple images with aligned camera locations can bring benefits to obtaining more accurate and robust depth estimation.

As shown in Fig.~\ref{fig:camera}, we command a robot arm to move a camera along its image plane so that all images are co-planar. Then the search for pixel correspondence can be conducted only on a fixed line direction. If we further move the camera along the horizontal or vertical axis, the disparity will only be along the horizontal or vertical axis. Furthermore, if the camera is moved with the same distance for every surrounding image, the disparity of each pixel relative to the center image should be the same, which is a strong regularization for computing an accurate disparity map. 

% what makes this work relevant
%Different from MVS which only do stereo correspondence matching in pairs, our system can treat three or more frames in stereo framework simultaneously because frames are co-planar and their camera centers are co-planar. This is difficult to form even with multiple cameras because the cameras cannot be identical and it is hard to adjust all cameras with co-planar image planes. And simultaneous rectification of more than two images is generally impossible \cite{szeliski2010computer} since their camera centers are not on one line.

%For stereo matching methods, the searching for pixel corresponding is hard because you only have one image to compare. The occlusion, reflection, illumination, no-texture and other challenges can influence the matching easily. But in a multiscopic system, even if the pixel in one surrounding frame cannot find a good matching with the center frame or the matching is wrong, it can be solved by the matching with other frames.

A multiscopic vision system brings clear benefits to depth estimation when compared to multiview stereo (MVS) and stereo matching. Compared to MVS that only performs stereo matching between pairs of images,\footnote{Note that simultaneous rectification of more than two images is generally impossible when their camera centers are not on one line \cite{szeliski2010computer}.} our system can aggregate cost volumes from all images simultaneously because all the captured images are aligned horizontally or vertically. For stereo matching, finding pixel correspondences is challenging because occlusion, reflection, illumination, no-texture can affect the matching easily. In a multiscopic system, depth estimation is much more robust in the presence of multiple cost volumes that can be combined. 

Previous approaches estimating disparity with multiscopic images fuse the information from different images with different types of heuristic terms \cite{kolmogorov2002multi, wei2005asymmetrical, drouin2005geo, maitre2008symmetric, lee2014multi}. These heuristic terms are based on hand-crafted prior knowledge and are usually overly simplified. In this paper, we first propose a new heuristic method to fuse multiple cost volumes to get a better disparity estimation. Then we propose a learning-based framework that trains a convolutional neural network called MFuseNet for cost volume fusion. The network is optimized to output the optimal disparity map by taking multiple cost volumes as input.  %The fusion is learned in the training by itself. There is no human design and the network operation could be much more complex. 

While there is no large multiscopic dataset for training, we use a 3D rendering engine to generate a synthetic multiscopic dataset with hundreds of different scenes. Although our network is trained with a small amount of data from this synthetic dataset, the model can be well generalized to real-world data and outperform heuristic methods. Our network is lightweight, with only tens of thousands of parameters, which makes the network well generalized to different scenes without overfitting. Besides, multiscopic matching with multiple aligned images generates much more accurate depth maps than two-view stereo matching. Visually, the depth map produced by multiscopic vision contains fairly few occlusion pixels, as shown in Fig.~\ref{fig:camera}. Note that in multiscopic vision, each pixel in the reference image likely appears in at least one of its surrounding images.

Our main contributions concerning regulated monocular perception system and multiscopic fusion are summarized as follows:

\begin{enumerate}
\item
We design a regulated perception system for accurate depth estimation. The system captures multiscopic images with the co-planar, parallel, and same-parallax structure using a monocular camera.
\item
%We propose a new heuristic algorithm and a neural network-based algorithm to do cost volume fusion and get better results than previous methods. With only a small number of synthetic data, the lightweight network could be trained easily and generalized well to real scenes.
We propose a novel learning-based model for multiscopic cost volume fusion. The proposed model obtains more accurate depth maps than previous methods do.
\item
%To train this network, we generate a synthetic multiscopic dataset with a 3D rendering engine, which could be used for more applications.
We generate a synthetic multiscopic dataset using a 3D rendering engine for training. With only tens of images, our lightweight network could be trained and generalized well to real-world scenes.
\end{enumerate}

% !TEX root =  ../main.tex

\section{RELATED WORK}
\label{sec:related_work}

We first review prior systems designed for capturing multiscopic images for depth estimation and afterward discuss algorithms that fuse multiscopic information to compute disparity maps.

%Active perception is widely employed in robotic applications such as exploration and manipulation \cite{chen2011active, bajcsy2018revisiting}. Active movement can assist in the localization of the manipulated objects under occlusions \cite{kahn2015active} or explore an unseen environment better \cite{isler2016information}.
%For stereo vision, since the baseline is critical for correspondence matching, some works about actively adjusting the baseline were proposed \cite{klarquist1997adaptive, nakabo2005variable}. A linear slider was used in \cite{nakabo2005variable} to change the baseline of two stereo cameras such that the baseline could be adaptive to the distance between the camera and the environment. This enables better 3D reconstruction of different scenes.

One type of multiscopic systems is based on camera arrays in which multiple cameras are placed on arrays \cite{wilburn2005high, vaish2006reconstructing, maitre2008symmetric}. A camera array brings the benefits that multiple cost volumes can be constructed by stereo matching between any two cameras on a row for robust depth estimation. It also resolves the partial invisibility issue as each point in the scene is likely visible in several cameras \cite{vaish2006reconstructing, maitre2008symmetric}. However, building a camera array with multiple cameras is bulky and expensive, and the rectification of different cameras is another challenge.

To take advantage of identical camera parameters, some stereo vision systems use a single camera to perform depth estimation. By analyzing the optical structure, Adelson and Wang proposed a single lens stereo system with a plenoptic camera that could produce photos from different viewpoints \cite{adelson1992single}. These captured images could be then used as stereo images for depth estimation. However, the stereo baseline is usually limited to the size of the lens aperture. Similar works using plates or mirrors to guide the light were proposed to obtain virtual stereo images. These optics design also introduces complex optical uncertainty and geometric calculation \cite{nene1998stereo, gao2004single, gluckman2002rectified, hu2017monocular}. All these systems are only for stereo vision rather than multiscopic vision.

%Since there is no many multiscopic systems, 
Most approaches for multiscopic vision are only demonstrated on synthetic data, and a few are performed on camera arrays.  
Kolmogorov et al. and Maitre et al. 
\cite{kolmogorov2002multi, maitre2008symmetric} treated the input images symmetrically while Wei et al. \cite{wei2005asymmetrical} treated the images asymmetrically. Occlusion models were built by Wei et al. \cite{wei2005asymmetrical} and Drouin et al. \cite{drouin2005geo} to handle the occlusion problem. Total variation regularization for multiple disparities and cross-filling was proposed for array disparity estimation \cite{lee2014multi}. Also, some methods calculate each disparity map with stereo matching and then merge the output disparity maps \cite{fehrman2014depth}.

All prior methods for the fusion of multiple cost volumes are manually designed, which may introduce human prior knowledge and bias. In this paper, we use a low-cost monocular camera to capture images in horizontally or vertically aligned camera positions. Then we propose a new heuristic fusion method and a deep learning-based fusion method to merge multiple cost volumes. To the best of our knowledge, this is the first time of fusing multiscopic cost volumes utilizing convolutional neural networks.

% !TEX root =  ../main.tex

\begin{figure}[]
\centering
\includegraphics[width=1.0\columnwidth, trim={0cm 0cm 0cm 0cm}, clip]{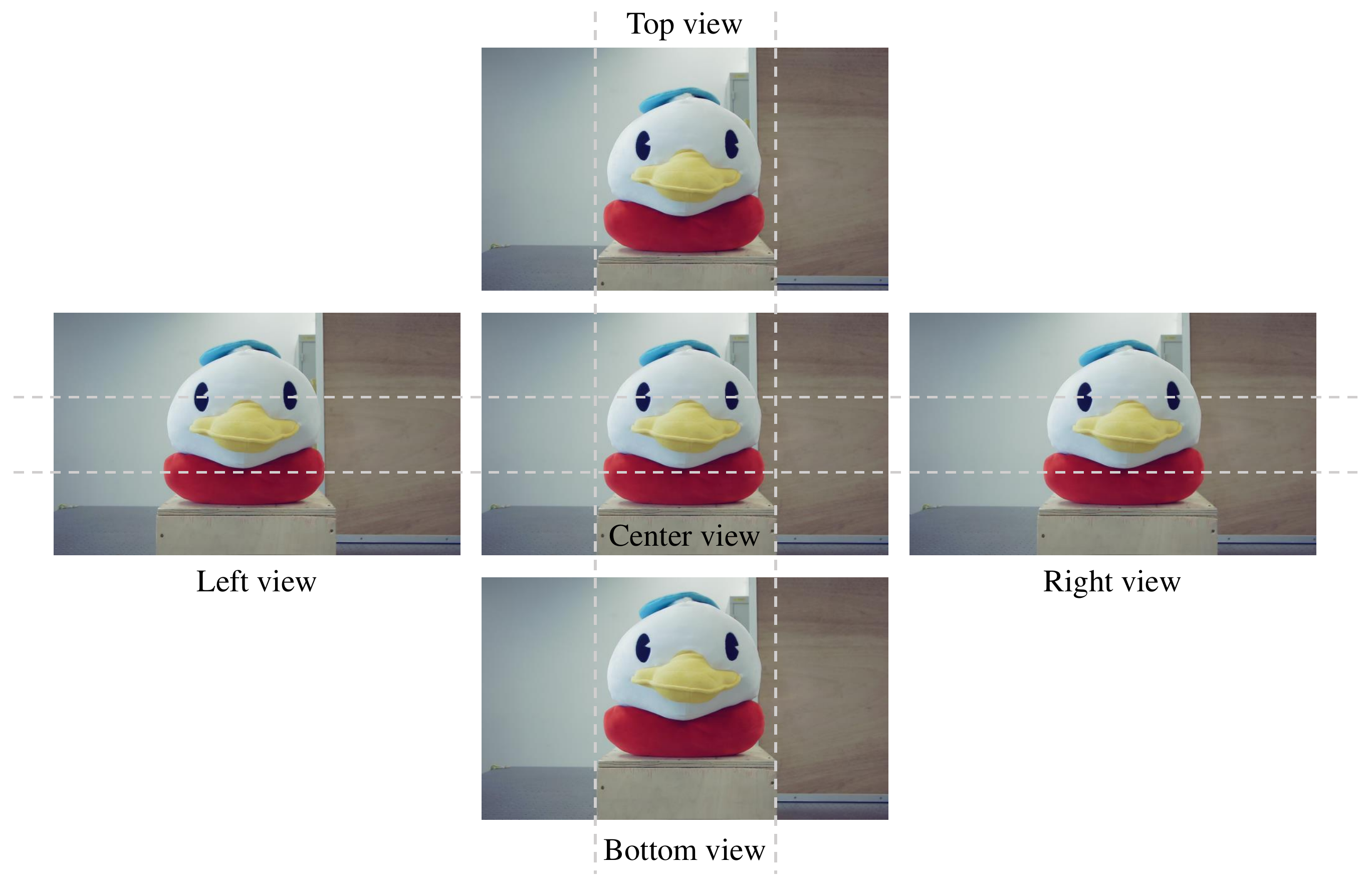}
\caption{Five images captured using our multiscopic perception system from different viewpoints. The parallax between the center view and any adjacent view is the same.}
\label{fig:5frame}
\vspace{-0.5cm}
\end{figure}

\section{HEURISTIC MULTISCOPIC FUSION}
\label{sec:multiscopic}

In this section, we first introduce our monocular multiscopic vision system to capture axis-aligned images and then propose our heuristic multiscopic fusion methods based on classical stereo algorithms to illustrate how the multiscopic matching is different from stereo matching. This can give us a hint about the role of MFuseNet and what is done inside the network.

\subsection{Monocular Multiscopic Perception}

% With only two frames, there are sometimes bad matches or noise. But if we have more frames this can be improved. 

% In our multiscopic system presented in Fig.~\ref{fig:camera}, we can move the monocular camera to the left and to the right along the horizontal, and move the camera up and down along the vertical, which generates four surrounding frames and one center frame, as displayed in Fig.~\ref{fig:5frame}. All frames are flat co-planar and the surrounding frames are with the same parallax 20 mm relative to the center frame. With the center frame as the reference, the other four frames can provide information and contribute to the disparity estimation together, which can suppress noise and ignore the bad matches. Besides, for occluded areas in one view, there is always other views that can perceive it. For example, the point $P$ in Fig.~\ref{fig:camera} cannot be observed from the left view, but can be perceived completely from other views.
In our multiscopic vision system presented in Fig.~\ref{fig:camera}, we can actively control the motion of a camera. We move it to the left and right along the horizontal axis, as well as up and down along the vertical axis. We capture one center image and four axis-aligned images with the same baseline in the left, right, bottom, and top views, as displayed in Fig.~\ref{fig:5frame}. 
The center image and any surrounding image can form a pair of stereo images.  
The five images can be captured with arbitrary baselines in our active perception system. The baseline can be adjusted for different purposes: accurate depth estimation for distant objects may require large baselines and stereo matching is easier with smaller baselines.% (less occlusion and smaller disparity).
%The baseline between the center image and one neighboring image is 20 millimeters. 

With the center image as the reference, the other four images can jointly contribute to the disparity estimation. Besides, each point in the center image is likely seen in one of the other four images. For example, the point $P$ in Fig.~\ref{fig:camera} cannot be observed from the left view but can be perceived completely from other views.

\subsection{Multiscopic Matching}

We will introduce block matching first and then multiscopic matching. Block matching is a simple and straightforward stereo matching method, which minimizes the matching error between two blocks in the left image and the right image. To find the most similar block, we need to check all possible blocks in the same row from the minimum disparity to the maximum plausible disparity. The sum of absolute difference (SAD) is often used to measure the visual similarity between two blocks. For a pixel $(u, v)$ in the left image, its SAD cost with block size $2\rho+1$ and  disparity $d$ can be calculated as

\begin{equation}
  c_{\text{SAD}}(u, v, d)=\sum_{x=u-\rho}^{u+\rho}\sum_{y=v-\rho}^{v+\rho}|I_l(x,y) - I_r(x-d, y) |,
\end{equation}
where $c_{\text{SAD}}(u,v,d)$ is the block matching cost at pixel $(u,v)$, $\rho$ is the radius of the block, $I_l(x,y)$ is the intensity of  pixel $(x,y)$ in the left image, and $I_r(x-d,y)$ denotes the intensity in the right image. 
%The center of the reference block is $(u, v)$ and the total number of pixels within this block is $(2\rho+1)^2$.

\begin{figure}[]
\centering
  \includegraphics[width=0.8\columnwidth, trim={0cm 0cm 0cm 0cm}, clip]{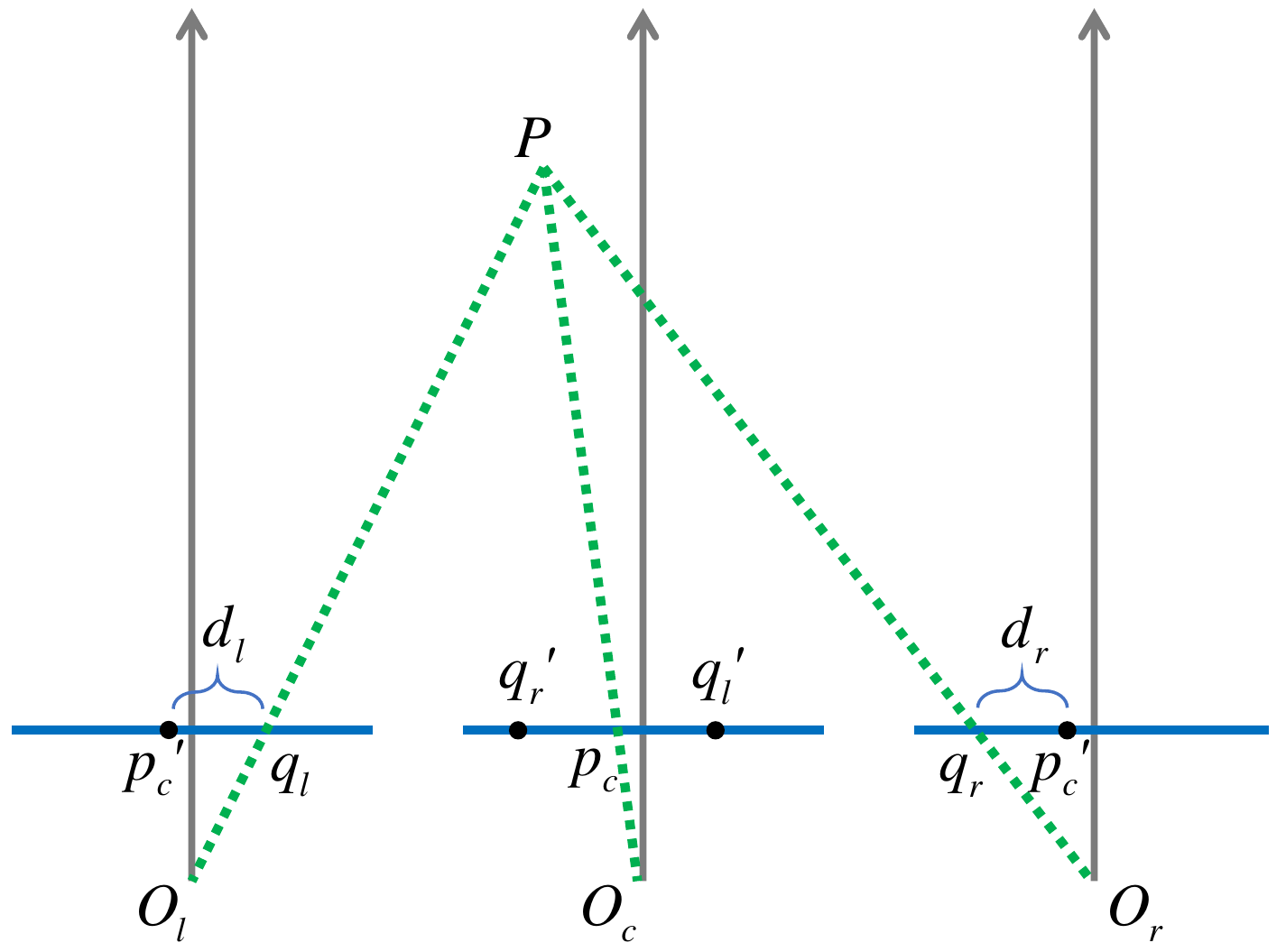}
\caption{A multiscopic system with three images is formed by moving a camera horizontally along the image plane with the same distance. Thus there are three images captured from the left view, the center view, and the right view. The gray optical axes are perpendicular to the image planes in blue. The points $O_l$, $O_c$, and $O_r$ are optical centers. A point $P$ in 3D space is projected onto the 2D image planes corresponding to the three pixels $q_l$, $p_c$, and $q_r$. $q_l'$, $q_r'$, and $q_c'$ are the counterparts of $q_l$, $q_r$, and $q_c$ in other images.}
\label{fig:optical}
\vspace{-0.3cm}
\end{figure}

The images in multiscopic vision are taken with parallel optical axes and co-planar image planes. Since the baselines for four surrounding images are the same, the disparity of a pixel between the center image and any surrounding image should be the same. 
This is demonstrated in Fig.~\ref{fig:optical}. Considering a multiscopic system with three images as an example, for a point $P$ in 3D space, it is projected onto the camera image planes as three image pixels $q_l$, $p_c$, and $q_r$. The disparity $d_l$ between $p_c$ and $q_l$ and the disparity $d_r$ between $p_c$ and $q_r$ are the same.

In real-world applications, our multiscopic vision system takes five images, as shown in Fig.~\ref{fig:5frame}. Thus the block matching cost is composed of four parts, each for one surrounding image: 
\begin{equation}
  \begin{aligned}
  c_{\text{SAD1}}(u,v,d)=\sum_{x=u-\rho}^{u+\rho}\sum_{y=v-\rho}^{v+\rho}|I_r(x-d, y)-I_c(x,y) |, \\
%   c_{\text{SAD2}}(u,v,d)=\sum_{x=u-\rho}^{x=u+\rho}\sum_{y=v-\rho}^{y=v+\rho}|I_l(x+d, y)-I_c(x,y) |, \\
%   c_{\text{SAD3}}(u,v,d)=\sum_{x=u-\rho}^{x=u+\rho}\sum_{y=v-\rho}^{y=v+\rho}|I_t(x, y+d)-I_c(x,y) |, \\
%   c_{\text{SAD4}}(u,v,d)=\sum_{x=u-\rho}^{x=u+\rho}\sum_{y=v-\rho}^{y=v+\rho}|I_b(x, y-d)-I_c(x,y) |,
  \end{aligned}
\end{equation}
where $I_r, I_c$ denote the images taken from the right view and the center view. $c_{\text{SAD2}}, c_{\text{SAD3}}, c_{\text{SAD4}}$ can be computed in a similar way between $I_c(x,y)$ and $I_l(x+d, y), I_t(x, y+d), I_b(x, y-d)$ in the left, top and bottom images, respectively. 
% where $I_r,I_l,I_t,I_b$ denote the images taken from the right, left, top, and bottom views respectively.
Then the fusion of these four cost functions to form the final data term is crucial. One naive idea is to take the average,
\begin{equation}
  c_{\text{ave}}=\frac{1}{4}(c_{\text{SAD1}}+c_{\text{SAD2}}+c_{\text{SAD3}}+c_{\text{SAD4}}).
\end{equation}

\begin{figure}[]
\centering
\begin{subfigure}{0.45\columnwidth}
  \centering
  \includegraphics[width=1\columnwidth, trim={2cm 2cm 4cm 1cm}, clip]{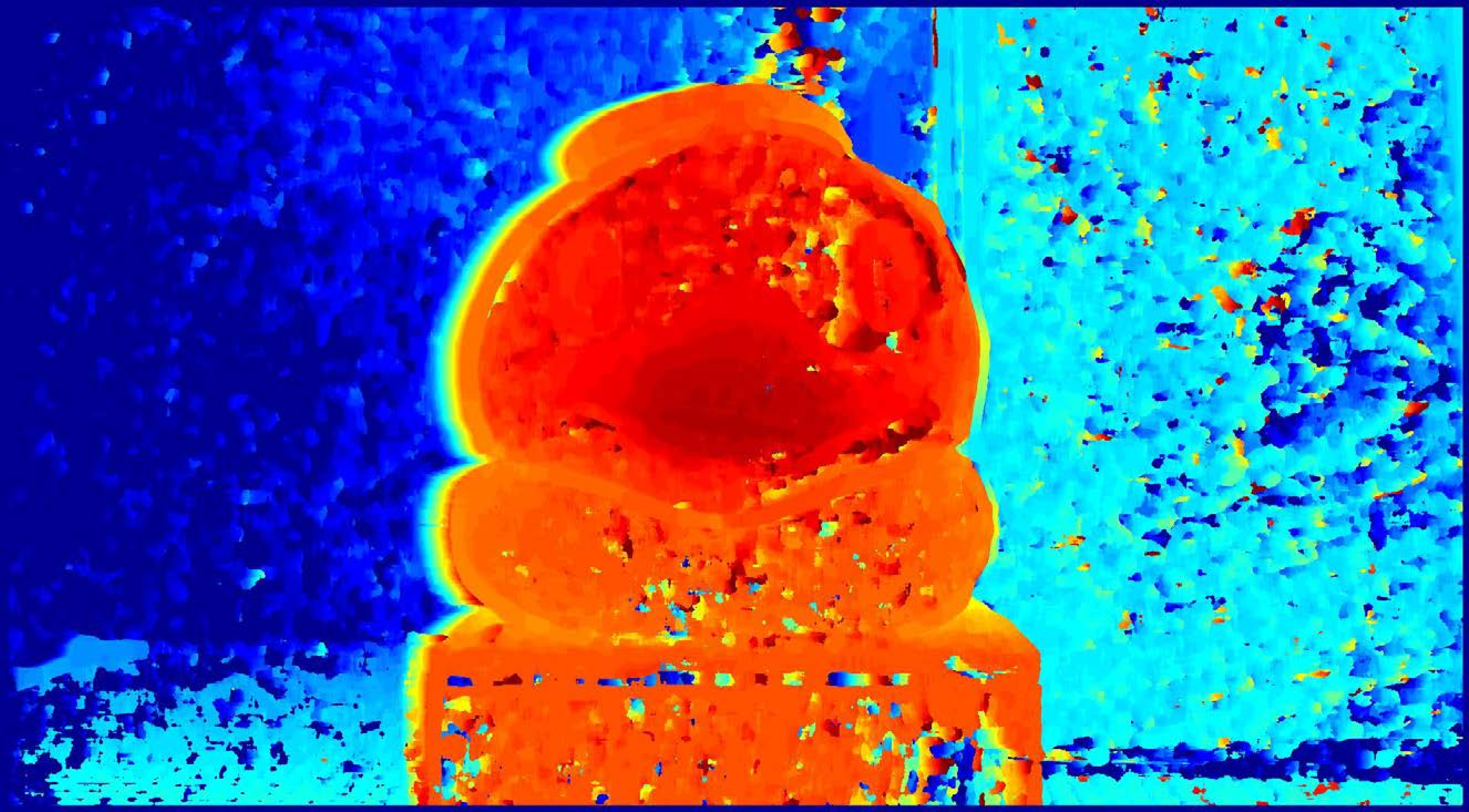}
  \caption{Stereo BM}
\end{subfigure}
\begin{subfigure}{0.45\columnwidth}
  \centering
  \includegraphics[width=1\columnwidth, trim={2cm 2cm 4cm 1cm}, clip]{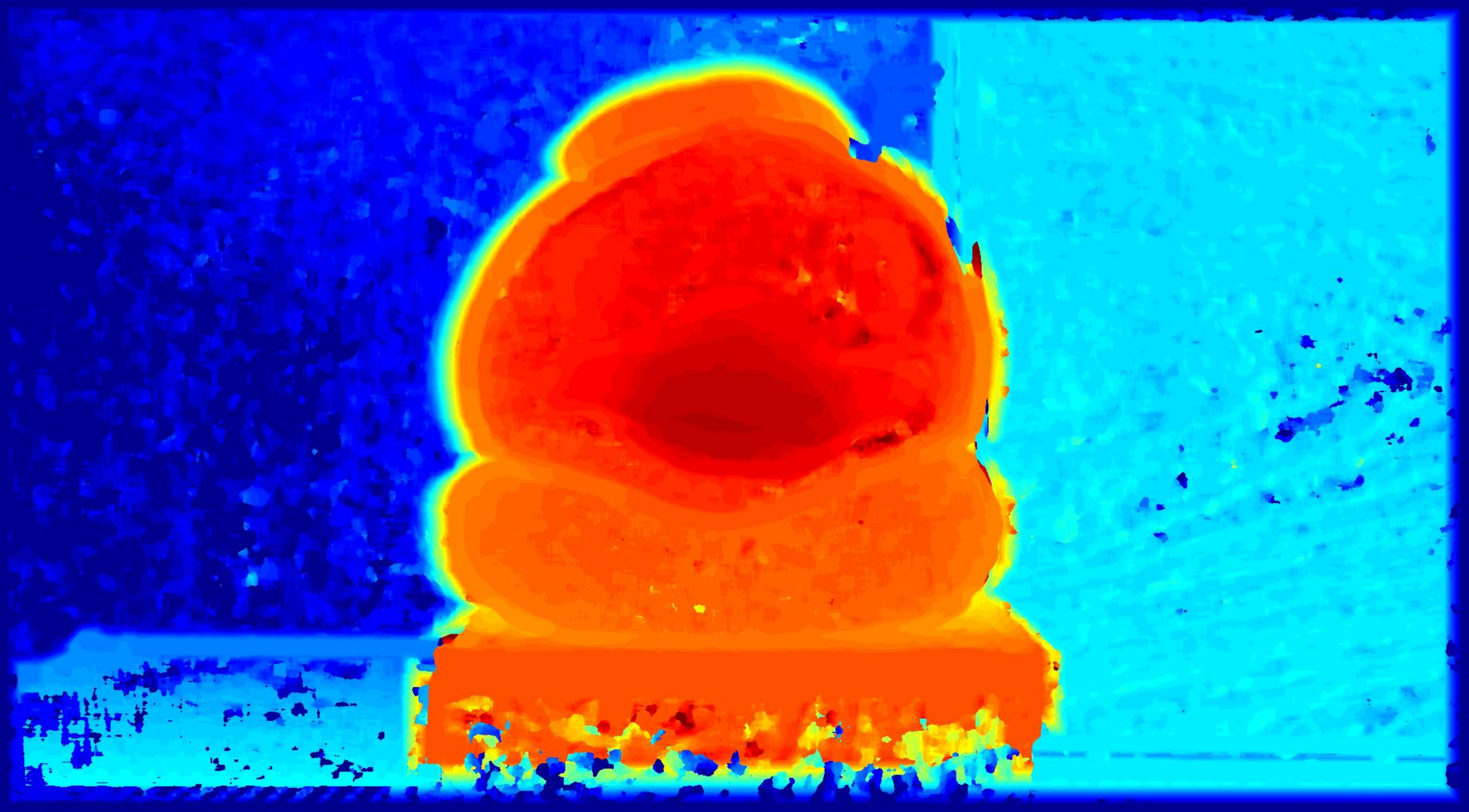}
  \caption{Mean fusion}
\end{subfigure}
\begin{subfigure}{0.45\columnwidth}
  \centering
  \includegraphics[width=1\columnwidth, trim={2cm 2cm 4Cm 1cm}, clip]{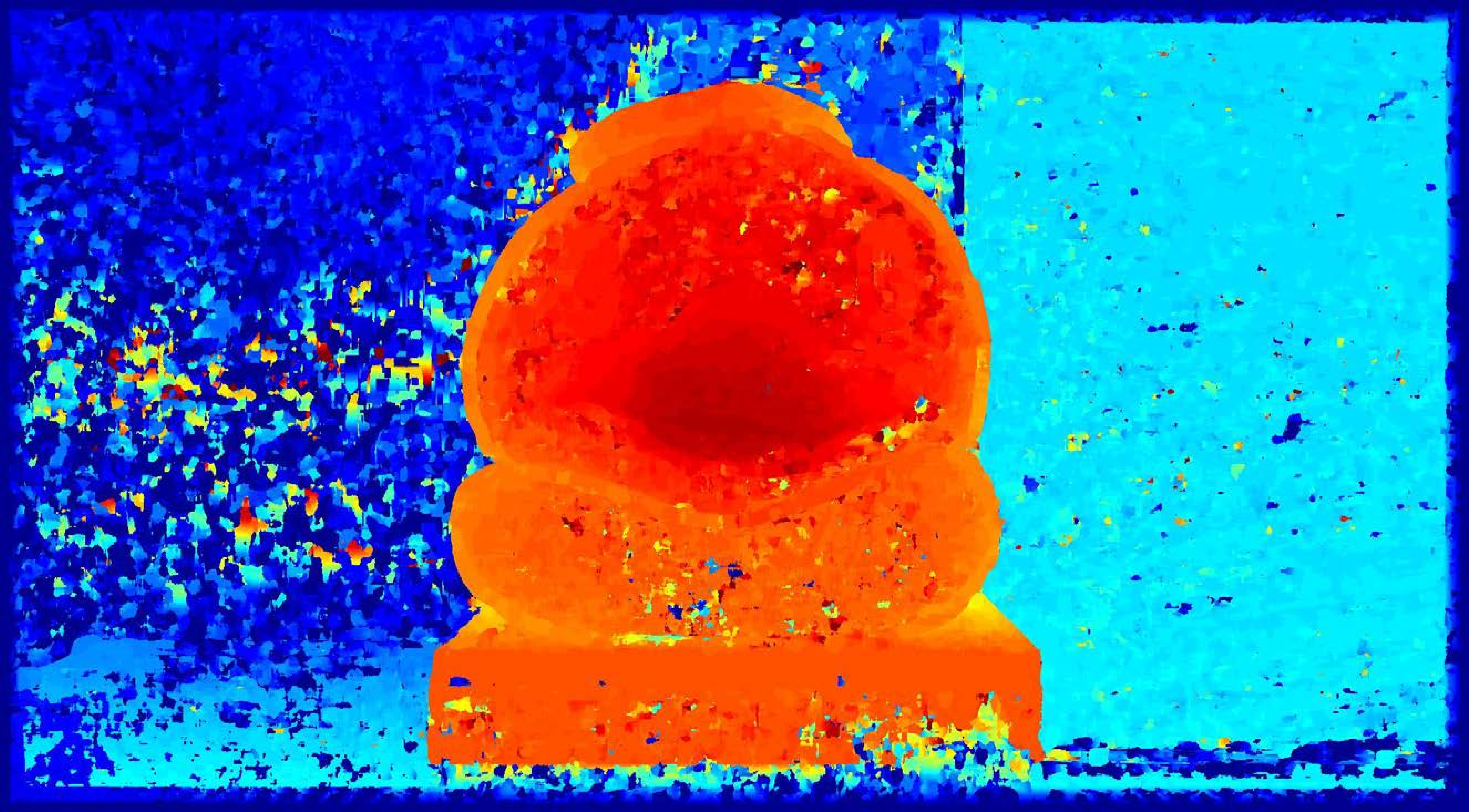}
  \caption{Minimum fusion}
\end{subfigure}
\begin{subfigure}{0.45\columnwidth}
  \centering
  \includegraphics[width=1\columnwidth, trim={2cm 2cm 4cm 1cm}, clip]{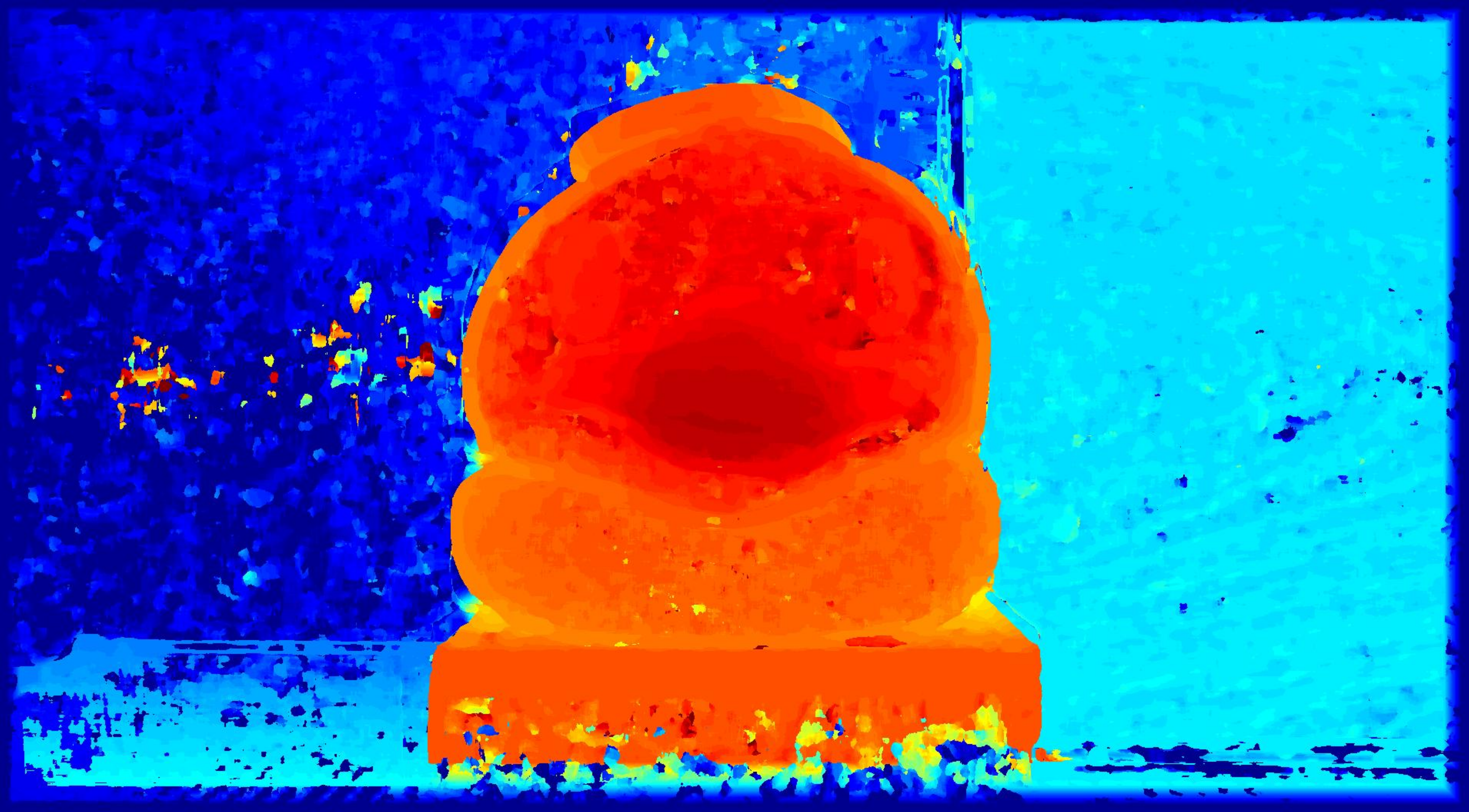}
  \caption{Heuristic fusion}
\end{subfigure}
% \begin{subfigure}{0.49\columnwidth}
%   \centering
%   \includegraphics[width=1\columnwidth, trim={2cm 1cm 4cm 1cm}, clip]{figures/duck/GC55}
%   \caption{Multiscopic graph cuts}
% \end{subfigure}
\caption{The disparity maps of stereo and multiscopic block matching. (a) Stereo BM is the block matching method using $c_{\text{SAD1}}$. Block matching with mean, minimum, and heuristic SAD cost fusion produce disparity maps (b), (c), (d) respectively, displayed in Jet colormap. All these methods use a winner-take-all strategy for the disparity maps.}
\label{fig:multi}
\vspace{-0.5cm}
\end{figure}

The visual result of using $c_{\text{ave}}$ shown in Fig.~\ref{fig:multi}(b) suggests that it does remove much more noise and reconstruct the reflective tabletop better than using $c_{\text{SAD1}}$ as shown in Fig.~\ref{fig:multi}(a), but disparity errors still exist in occluded areas. For the center image, some regions can not be seen in some surrounding images. For instance, the region to the left of the toy cannot be seen in the right image. Thus the cost $c_{\text{SAD1}}$ for this region would be large and may affect the overall data term $c_{\text{ave}}$. Therefore we consider another fusion strategy by choosing the smallest one when combining the four cost functions:
\begin{equation}
  c_{\text{min}}=\min\{c_{\text{SAD1}},c_{\text{SAD2}},c_{\text{SAD3}},c_{\text{SAD4}}\}.
  \label{equ:min}
\end{equation}

The disparity map computed with $c_{\text{min}}$ is presented in Fig.~\ref{fig:multi}(c). We can see that the occlusion region is reconstructed clearly, but the noise is persistent in some areas. To overcome this, we design a heuristic fusion strategy. We first sort the four costs on each pixel and use three smallest costs $c^{\text{\Romannum{1}}}, c^{\text{\Romannum{2}}}, c^{\text{\Romannum{3}}}$ ($c^{\text{\Romannum{1}}}$ is the smallest). Then we remove the second-highest cost if it is much higher than the other two:
\begin{equation}
c_{\text{heu}}=
\begin{cases}
\frac{1}{2}(c^{\text{\Romannum{1}}}+c^{\text{\Romannum{2}}}), \quad & \text{if} \ c^{\text{\Romannum{3}}}>3 c^{\text{\Romannum{2}}}\\
\frac{1}{3}(c^{\text{\Romannum{1}}}+c^{\text{\Romannum{2}}}+c^{\text{\Romannum{3}}}), \quad & \text{otherwise}
\end{cases}
,
\label{equ:heuristic}
\end{equation}
which leads to a cleaner disparity map as shown in Fig.~\ref{fig:multi}(d).

\subsection{Multiscopic Graph Cuts}

Graph cuts (GC) optimization is one of the most popular global optimization methods for stereo matching. It is a process that assigns a label of disparity to each pixel in the reference image such that an energy function is minimized. Our multiscopic graph cuts model is based on the stereo matching algorithm by Kolmogorov and Zabih \cite{kolmogorov2014kolmogorov}.

In our graph cuts optimization, the energy is composed of 4 terms defined as
\begin{equation}
  E=E_{\text{data}}+E_{\text{occlusion}}+E_{\text{smooth}}+E_{\text{unique}} \ .
\end{equation}

\textbf{Data term} $E_{\text{data}}$ is used to evaluate the similarity of two image patches. Note that our images may not be perfectly aligned due to the limited precision of robot arm movement, the epipolar line may deviate slightly from the horizontal or vertical direction. To compensate this, we use an improved Birchfield and Tomasi's (BT) dissimilarity for the data term \cite{birchfield1998pixel, kolmogorov2014kolmogorov}:
\begin{equation}
\begin{aligned}
  c_{\text{BT}1}(u,v,d)=\max\{0, I_c(u,v)-I_r^{\min}(u-d,v), \\ I_r^{\max}(u-d,v)-I_c(u,v)\},\\
\end{aligned}
\end{equation}
where $I_r^{\min}$ and $I_r^{\max}$ are respectively the smallest and largest values on the subpixel neighborhood around pixel $(u-d, v)$ in the right image. For a pixel $q$ in the right image:
\begin{equation}
\begin{aligned}
  I_r^{\min}(q)=\min_\sigma\{\frac{1}{2}(I_r(q)+I_r(q+\sigma)) \},\\
  I_r^{\max}(q)=\max_\sigma\{\frac{1}{2}(I_r(q)+I_r(q+\sigma)) \},
\end{aligned}
\end{equation}
where $\sigma\in\{(0,0),(-1,0),(1,0),(0,-1),(0,1)\}$. Therefore the stereo matching for correspondence is actually performed between the half higher row and the half lower row.

$E_{\text{data}}$ now is an integration of four parts. The other three are between $I_c(u,v)$ and $I_l(u+d,v), I_t(u,v+d), I_b(u,v-d)$.
% \begin{equation}
% \begin{aligned}
% c_{\text{BT}2}(u,v,d)=\max\{0, I_c(u,v)-I_l^{\min}(u+d,v), \\ I_l^{\max}(u+d,v)-I_c(u,v)\},\\
% c_{\text{BT}3}(u,v,d)=\max\{0, I_c(u,v)-I_t^{\min}(u,v+d), \\ I_t^{\max}(u,v+d)-I_c(u,v)\},\\
% c_{\text{BT}4}(u,v,d)=\max\{0, I_c(u,v)-I_b^{\min}(u,v-d), \\ I_t^{\max}(u,v-d)-I_c(u,v)\},
% \end{aligned}
% \end{equation}
% where $I_l^{\min}, I_l^{\max}, I_t^{\min}, I_t^{\max}, I_b^{\min}, I_b^{\max}$ are the smallest and largest values on the subpixel neighborhood in the left, top, bottom image respectively. 
These four costs are then merged using the same heuristic rule to get the fused cost $c_{\text{GC}}(u,v,d)$.

\begin{figure}[]
\centering
\begin{subfigure}{0.45\columnwidth}
  \includegraphics[width=1\columnwidth, trim={2cm 1cm 4cm 1cm}, clip]{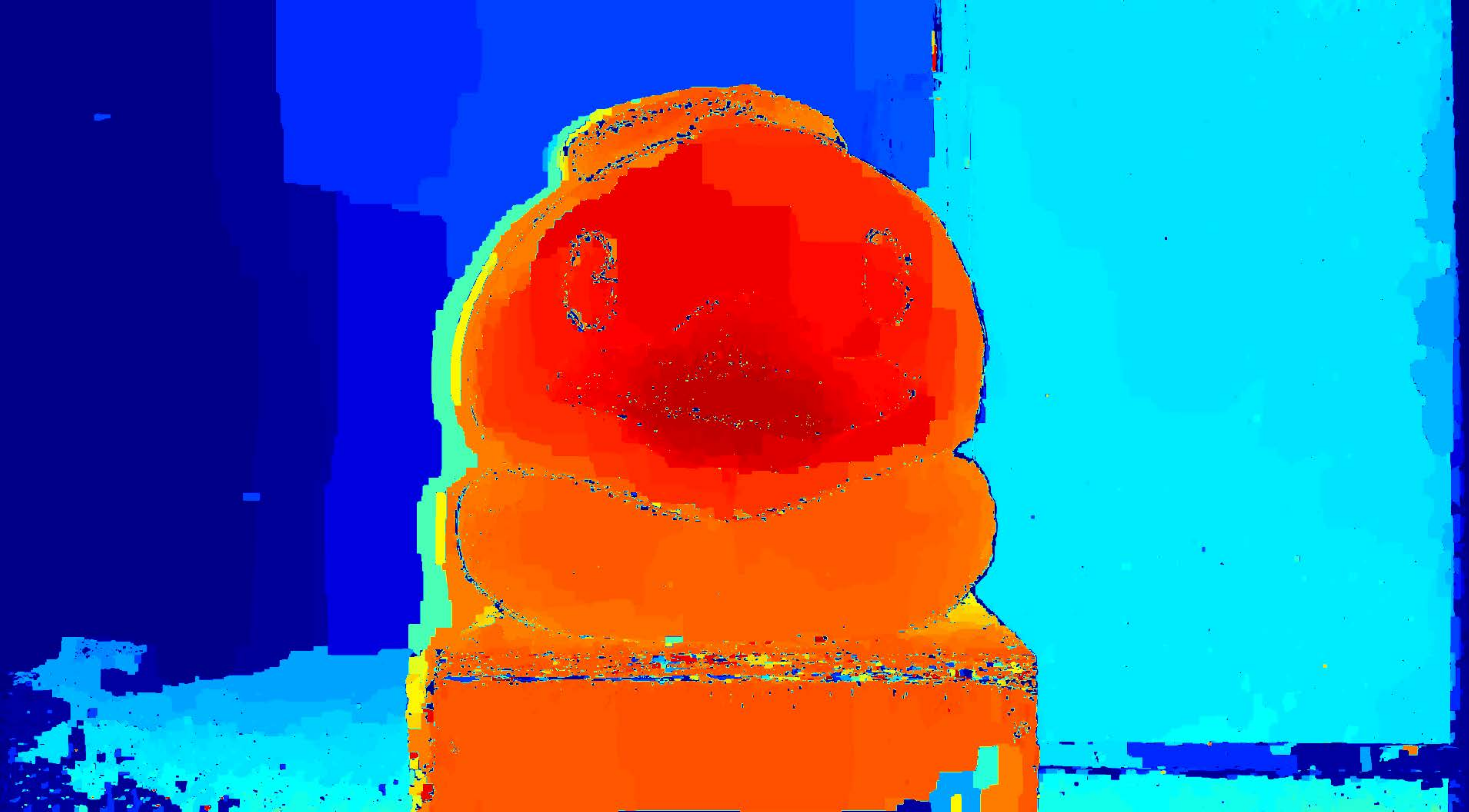}
  \caption{Stereo GC}
\end{subfigure}
\begin{subfigure}{0.45\columnwidth}
  \includegraphics[width=1\columnwidth, trim={2cm 1cm 4cm 1cm}, clip]{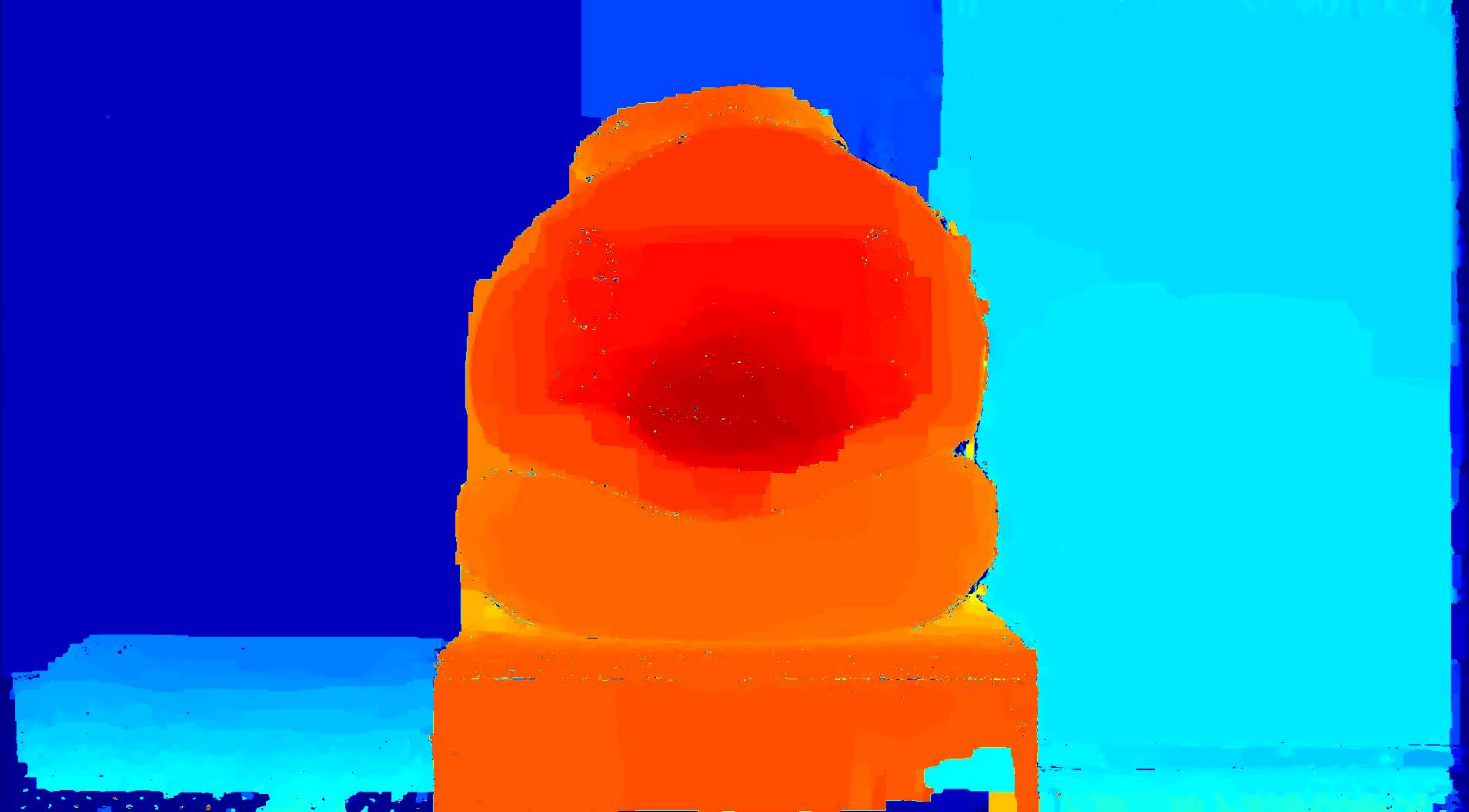}
  \caption{Heuristic Multiscopic GC}
\end{subfigure}
\caption{The disparity maps obtained by stereo and multiscopic graph cuts using heuristic fusion method. Note that for stereo graph cuts, the disparities on occluded regions are not estimated accurately, and matching on the metal tabletop is not accurate due to reflection.}
\label{fig:gc}
\vspace{-0.5cm}
\end{figure}

\textbf{Occlusion term} $E_{\text{occlusion}}$ is used to maximize the number of matches. To encourage more disparity assignment in graph cuts optimization, any inactive pixel without assignment is penalized by a constant $K$ \cite{kolmogorov2014kolmogorov}.

\textbf{Smoothness term} $E_{\text{smooth}}$ encourages assigning similar disparity to adjacent pixels, especially for those with a similar color. Thus if two adjacent pixels $p_1, p_2$ in the center image have different disparity assignments, a $L_1$ penalty would be added as:
\begin{equation}
\begin{split}
  V&=
  \begin{cases}
    \lambda_1 \Delta d, \quad \max\{|I_c(p_1)-I_c(p_2)|, |I_r(q_1)-I_r(q_2)|,\\
    \qquad \qquad \ |I_l(m_1)-I_l(m_2)|,
    |I_t(n_1)-I_t(n_2)|,\\
    \qquad \qquad \ |I_b(k_1)-I_b(k_2)|
    \}<\theta \\
    \lambda_2 \Delta d, \quad \text{otherwise}
  \end{cases}\\
  \Delta d &=\min\{|d_1-d_2|,\ d_{\text{CUTOFF}}\},
\end{split}
\end{equation}
where $q_1, q_2, m_1, m_2, n_1, n_2, k_1, k_2$ are the corresponding pixels of $p_1,p_2$ in the right, left, top, bottom images respectively, $\theta$ is a threshold to evaluate the color similarity, $\lambda_1$ and $\lambda_2$ are penalty constants, and $\Delta d$ is the disparity difference truncated at a threshold $d_{\text{CUTOFF}}$.

\textbf{Uniqueness term} $E_{\text{unique}}$ enforces the uniqueness of pixel correspondences. For a pixel in the center image, we do not allow two pixels in the surrounding image match it simultaneously and vice versa. This will be punished by an infinity penalty of $\infty$ \cite{kolmogorov2014kolmogorov}.

With this energy, the visual results of multiscopic graph cuts and stereo graph cuts with the same hyper-parameters are displayed in Fig.~\ref{fig:gc}. To suppress the discontinuous disparity artifacts, we enlarge input images twice before the graph cuts optimization. Compared with the stereo graph cuts, the occlusion parts and reflective tabletop are reconstructed much better, and the noise is better suppressed.

% !TEX root =  ../main.tex

\section{MULTISCOPIC FUSION with MFUSENET}

Manually designed heuristic fusion approaches are prone to human bias and are usually simplified. To better fuse multiple cost volumes, we propose to use deep neural networks to merge them. In this section, we first present the network structure we use to fuse cost volumes, after which we describe how we generate the synthetic dataset.

\subsection{Network Structure}

\begin{figure}[]
\centering
  \includegraphics[width=1\columnwidth, trim={0cm 0cm 0cm 0cm}, clip]{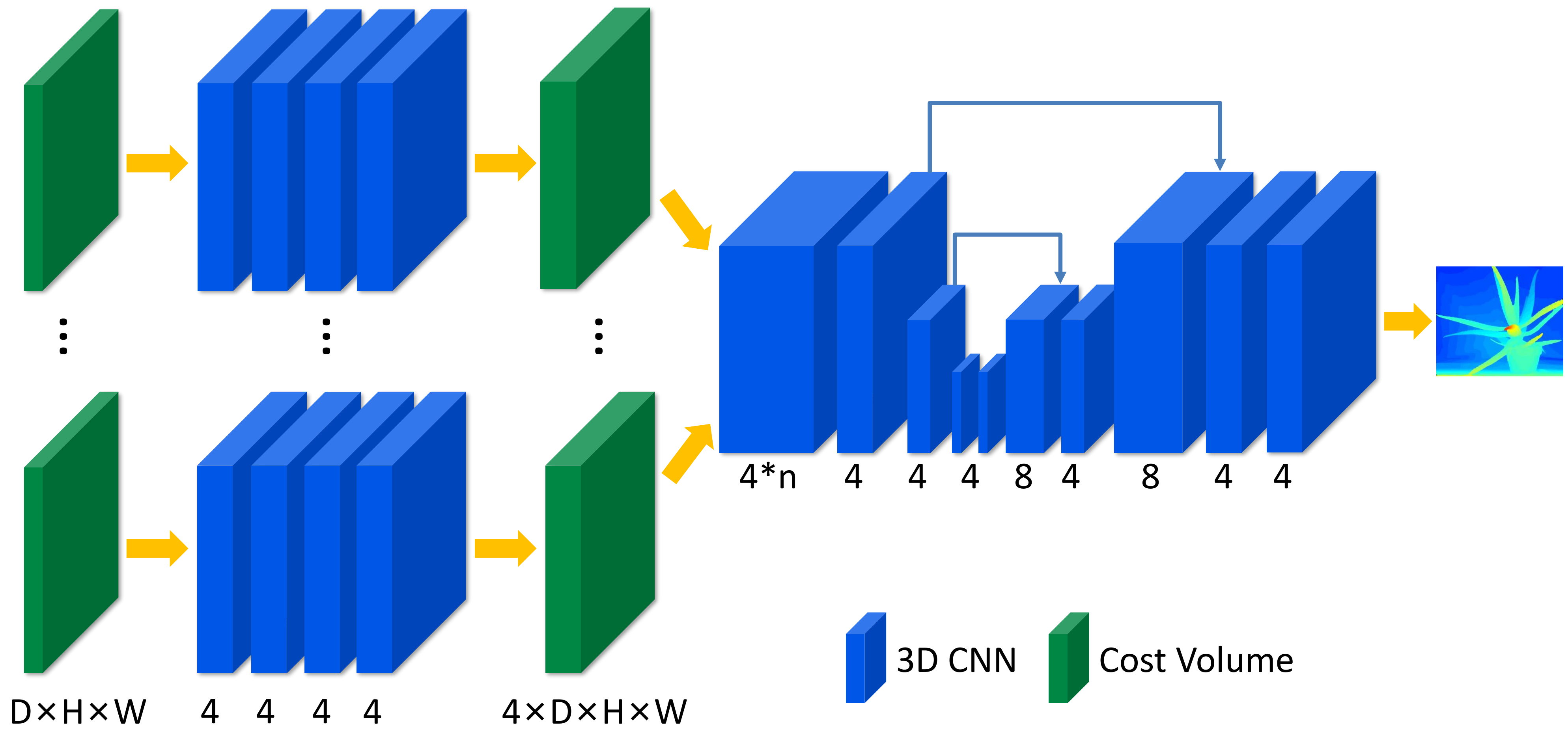}
\caption{The network structure of MFuseNet. For $n$ cost volumes with size $D\times H\times W$, they are processed respectively and then fused to get the final disparity. The feature channels of 3D CNN is 4 such that the size of each cost volume before concatenation is $4\times D\times H\times W$.}
\label{fig:network}
\vspace{-0.5cm}
\end{figure}

%We employ 3D Convolution Neural Network (3D CNN) to fuse multiple cost volumes. As is displayed in Fig.~\ref{fig:network}, the cost volumes are first fed into a cascaded 3D CNN layers to extract features and then concatenated together, after which a structure similar to U-Net \cite{ronneberger2015u} but in 3D level is utilized to fuse cost volumes. In this way different scale of information can be processed to incorporate more global information. Since the size of cost volumes are large, we only use small size of feature levels, in which case the number of parameters is only around 10 thousand. This makes our network easily trained and can be generalized to unseen scenarios well rather than over-fitting to the training data.  
We employ a 3D convolutional neural network (3D CNN) to fuse multiple cost volumes. As displayed in Fig.~\ref{fig:network}, each cost volume is first fed into cascaded 3D CNN layers to extract features. These features are concatenated as the input to a network similar to 3D U-net to generate the disparity prediction \cite{ronneberger2015u}. With the 3D U-net architecture, information at different scales can be processed to incorporate both local and global information. Since the size of cost volumes is large, we only use a small number of feature channels, such that the number of parameters in MFuseNet is only around 10 thousand. A compact network makes it easy to train and be generalized to unseen scenarios well rather than over-fitting to the training data.  

To obtain the final output, we apply a disparity regression function \cite{kendall2017end} at the end of the network to get a continuous disparity map. A softmax activation function is utilized to get a probability map with size $1\times D \times H \times W$. Then the final disparity is calculated as the sum of each disparity $d$ weighted by its probability $p_d$:
\begin{equation}
    \hat{d}_i = \sum_{d=1}^{D} d \times p_d \ .
\end{equation}

The final loss for a disparity map with $N$ pixel is calculated as the smooth $L_1$ loss:
\begin{equation}
L(\hat{d}, d^{gt}) = \frac{1}{N}\sum_{i=1}^N l(\hat{d}_i-d^{gt}_i) \ ,
\end{equation}
where $d^{gt}$ is the ground-truth disparity map and
\begin{equation*}
l(x)=
\begin{cases}
\frac{1}{2}x^2  \ &|x|<1   \\
|x|-\frac{1}{2}   \ &otherwise
\end{cases} \ .
\end{equation*}
With this loss, the fusion network can be trained with cost volumes calculated by any algorithm. In this paper we only use the ones obtained by MC-CNN \cite{zbontar2016stereo} since it generates intermediate cost volumes after semi-global matching.

\subsection{Synthetic Multiscopic Dataset}

\begin{figure}[]
\centering
\begin{subfigure}{0.3\columnwidth}
  \centering
  \includegraphics[width=1\columnwidth, trim={0cm 0cm 0cm 0cm}, clip]{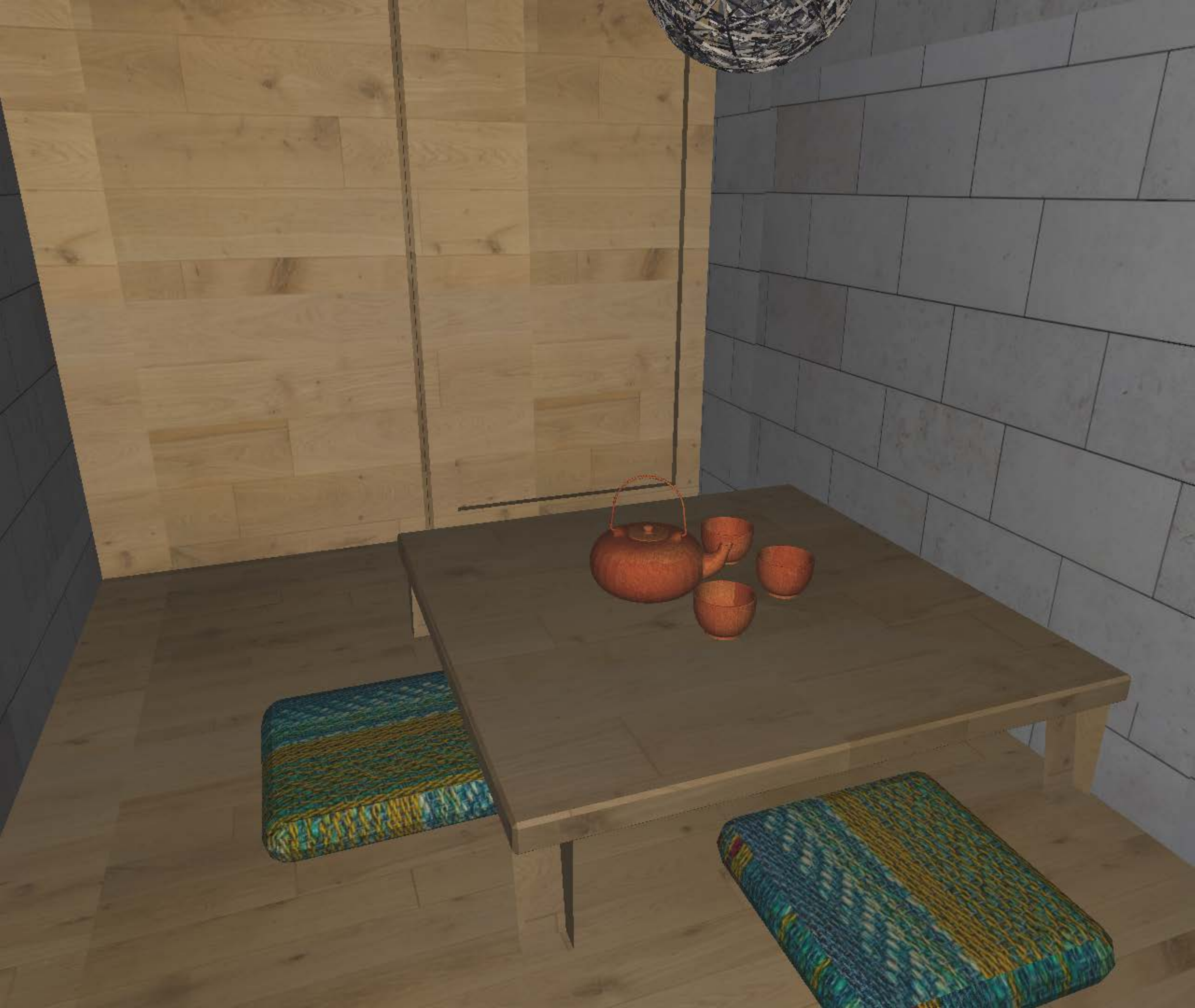}
  \caption{Synthetic image}
\end{subfigure}
\begin{subfigure}{0.3\columnwidth}
  \centering
  \includegraphics[width=1\columnwidth, trim={0cm 0cm 0cm 0cm}, clip]{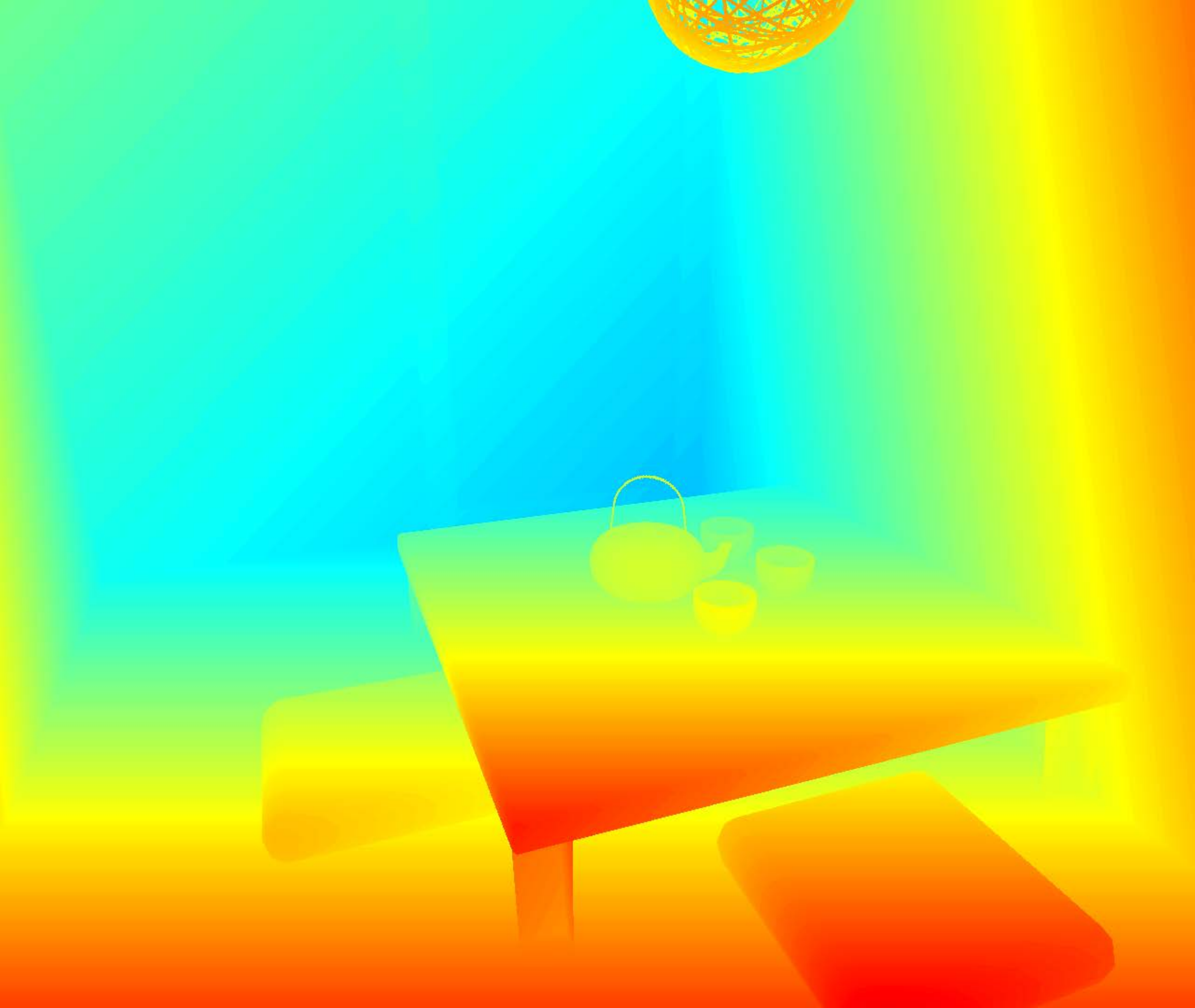}
  \caption{Ground truth}
\end{subfigure}
\begin{subfigure}{0.3\columnwidth}
  \centering
  \includegraphics[width=1\columnwidth, trim={0cm 0cm 0cm 0cm}, clip]{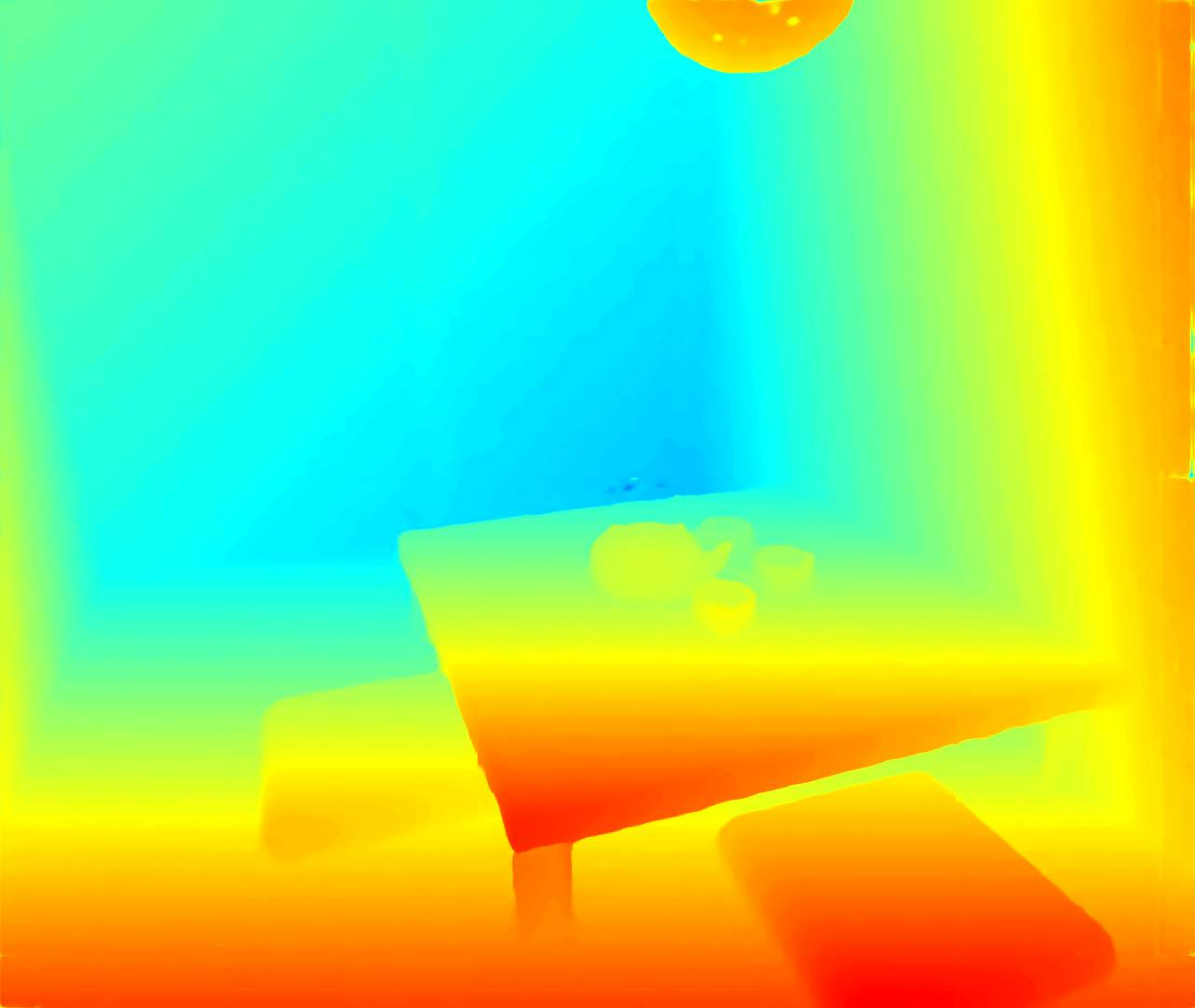}
  \caption{MFuseNet}
\end{subfigure}
\begin{subfigure}{0.3\columnwidth}
  \centering
  \includegraphics[width=1\columnwidth, trim={0cm 0cm 0cm 0cm}, clip]{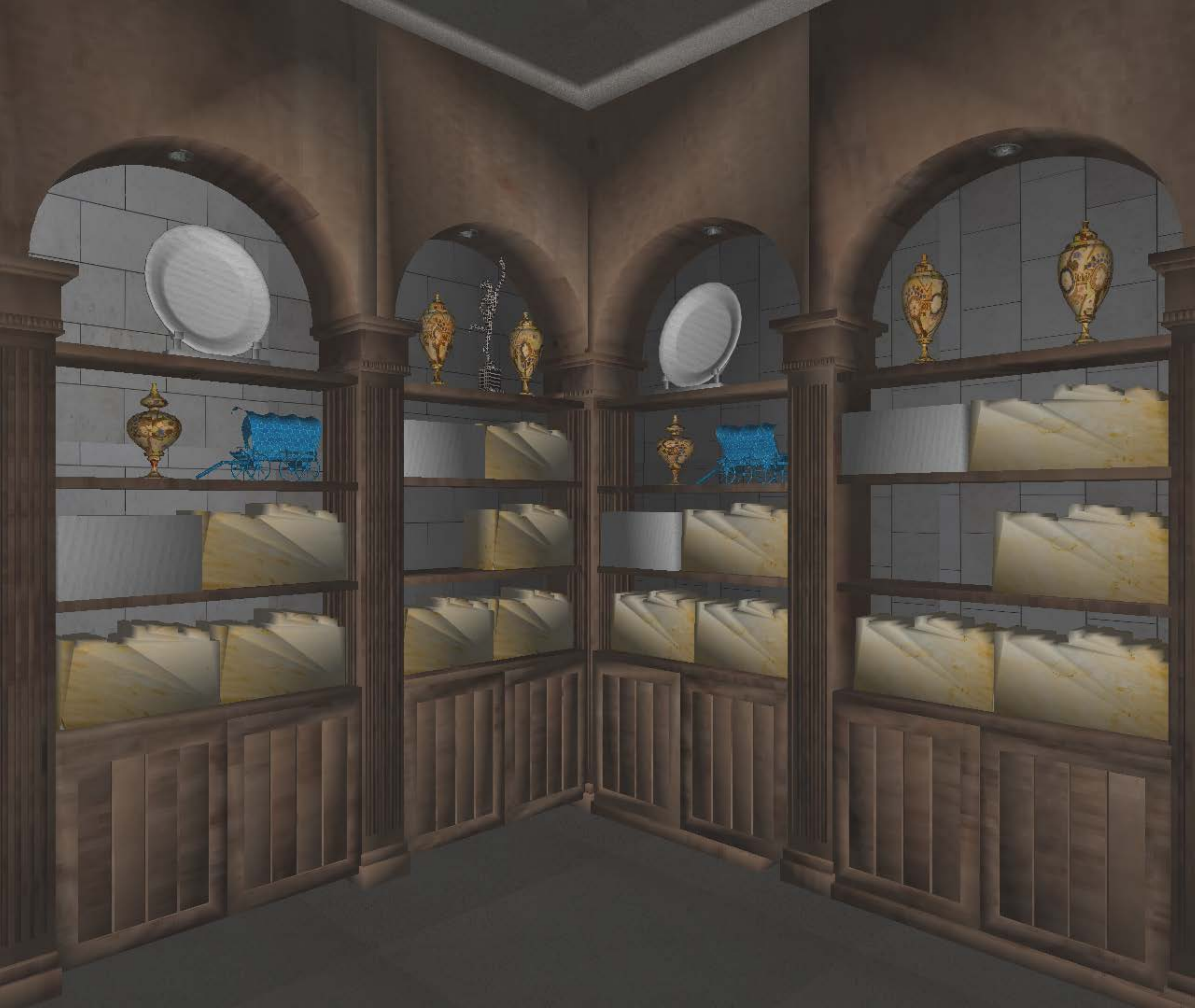}
  \caption{Synthetic image}
\end{subfigure}
\begin{subfigure}{0.3\columnwidth}
  \centering
  \includegraphics[width=1\columnwidth, trim={0cm 0cm 0cm 0cm}, clip]{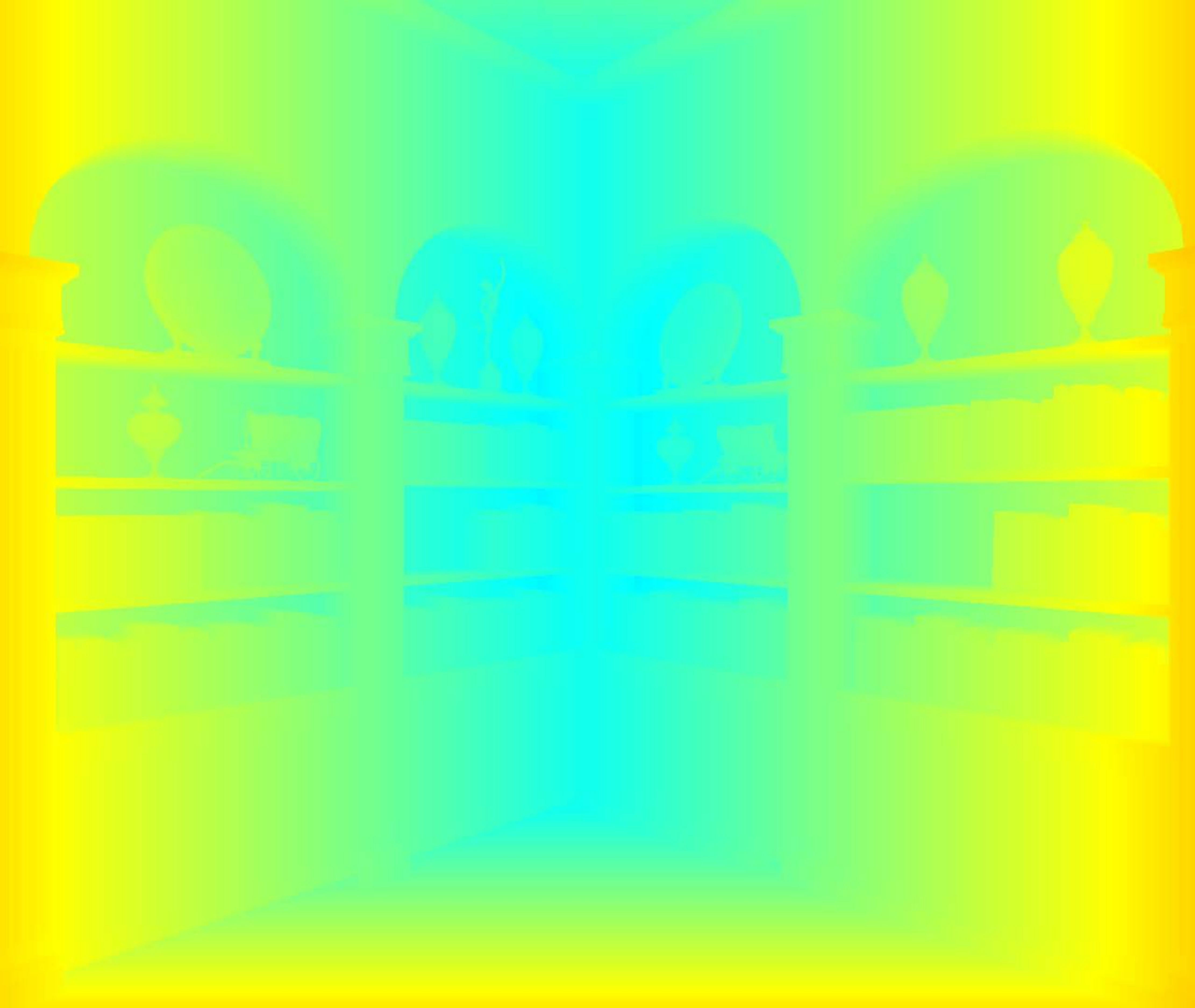}
  \caption{Ground truth}
\end{subfigure}
\begin{subfigure}{0.3\columnwidth}
  \centering
  \includegraphics[width=1\columnwidth, trim={0cm 0cm 0cm 0cm}, clip]{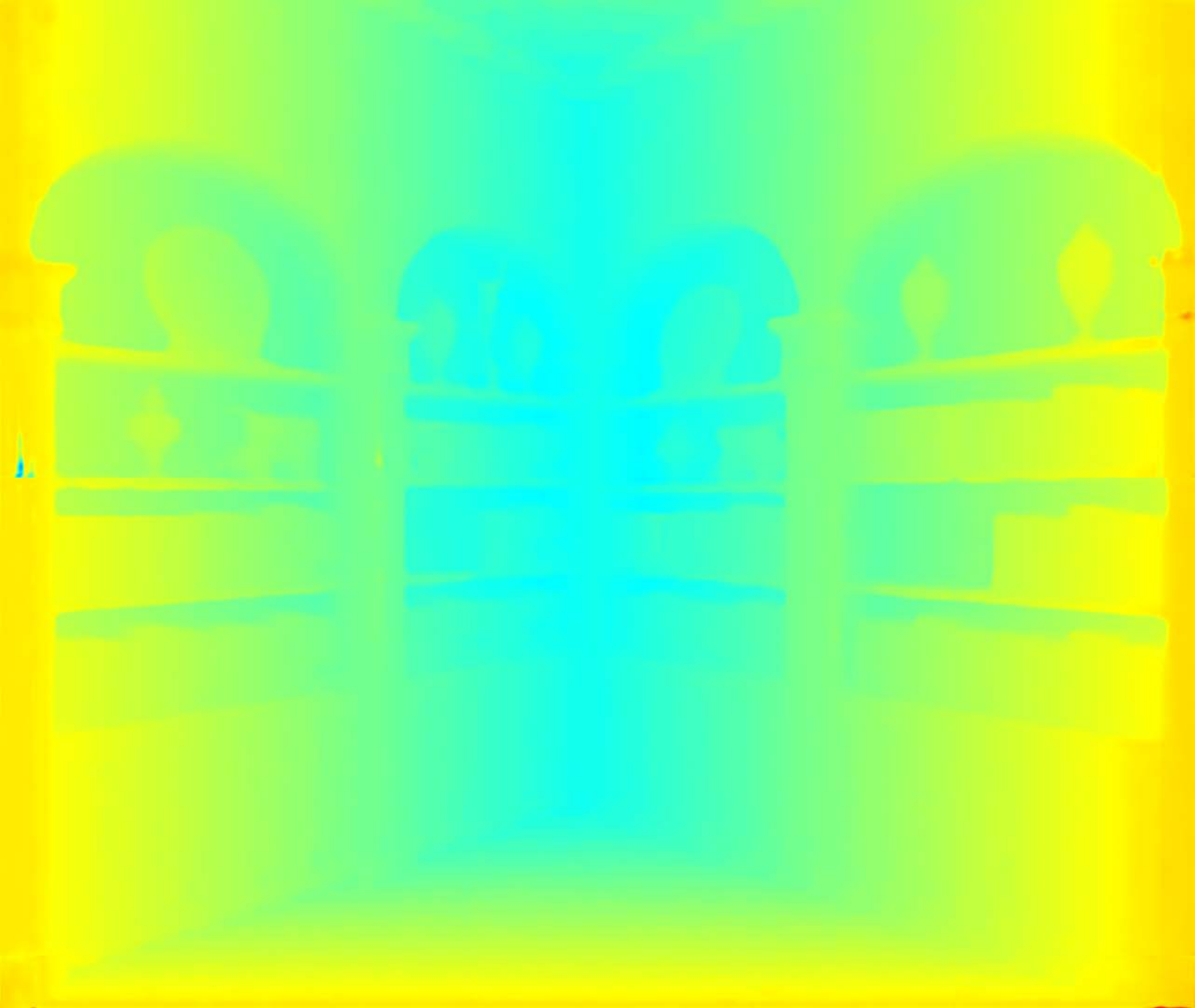}
  \caption{MFuseNet}
\end{subfigure}

\caption{Color images and ground-truth disparity maps in the synthetic multiscopic dataset,  and the disparity maps obtained by MFuseNet.}
\label{fig:synthetic}
% \vspace{-0.3cm}
\end{figure}

\begin{figure}[]
\centering
\begin{subfigure}{0.45\columnwidth}
  \centering
  \includegraphics[width=1\columnwidth, trim={0cm 0cm 0cm 0cm}, clip]{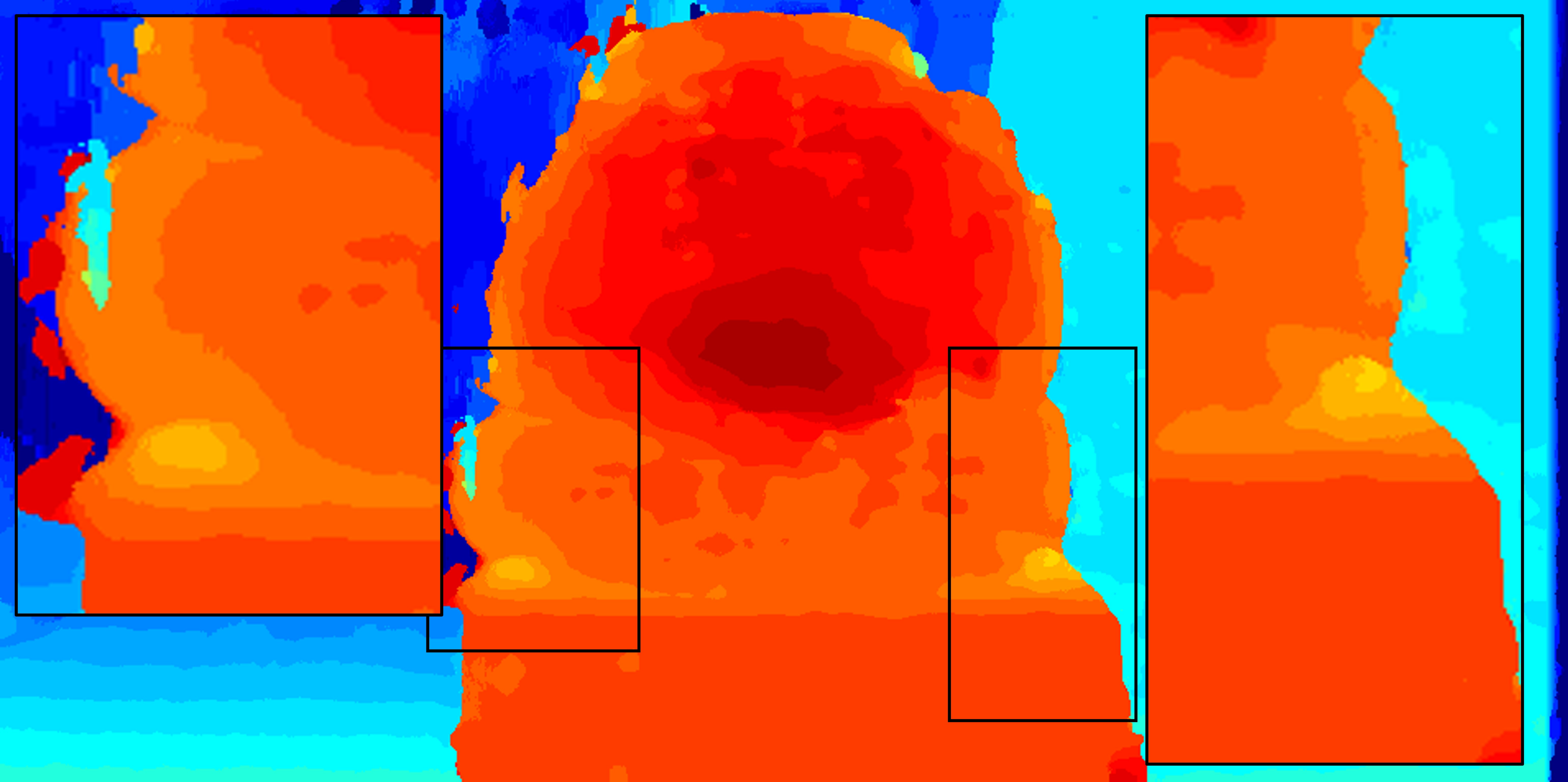}
  \caption{Stereo MC-CNN}
\end{subfigure}
\begin{subfigure}{0.45\columnwidth}
  \centering
  \includegraphics[width=1\columnwidth, trim={0cm 0cm 0cm 0cm}, clip]{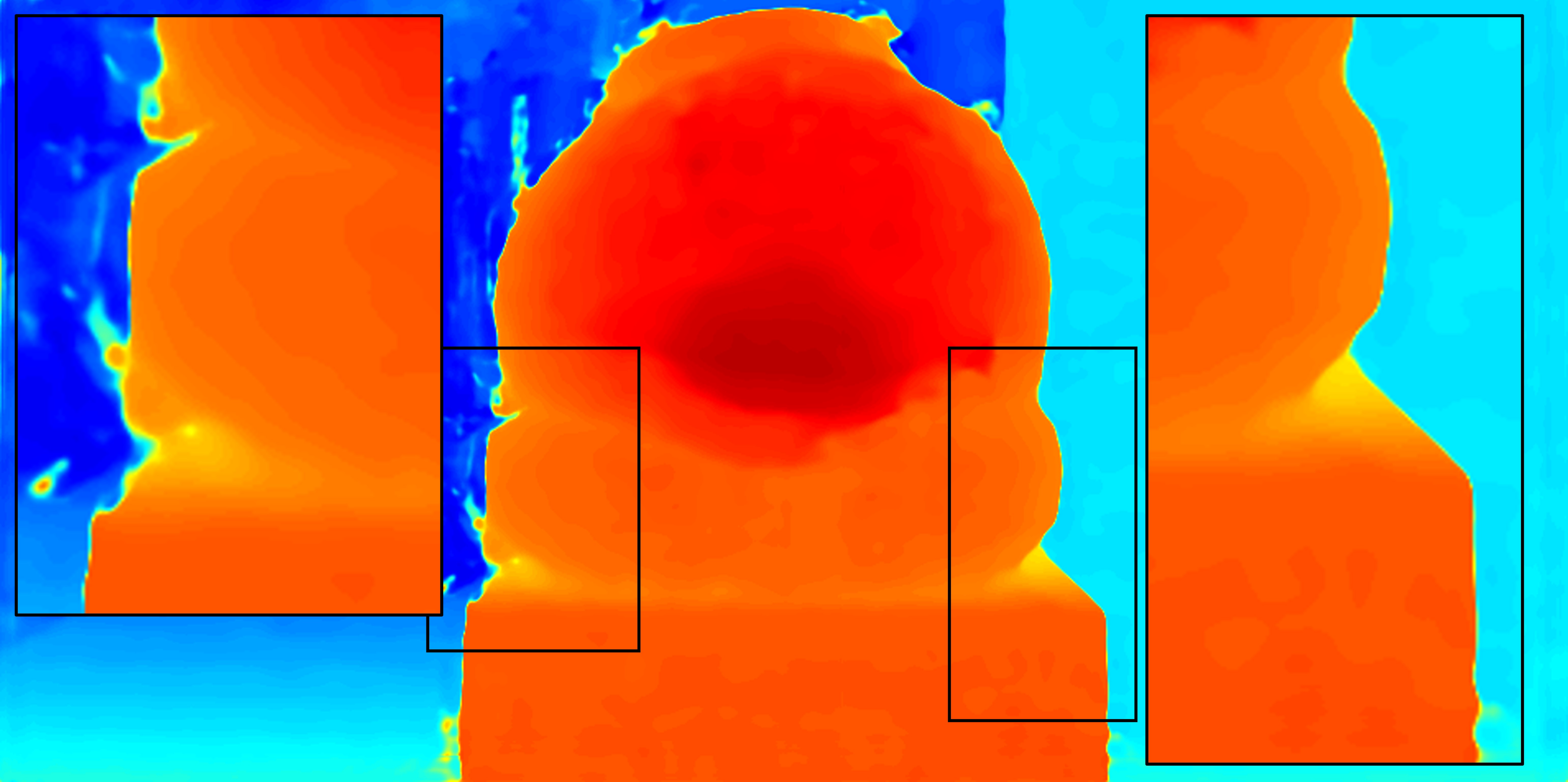}
  \caption{MFuseNet}
\end{subfigure}
\caption{The disparity maps obtained by MC-CNN using stereo images and MFuseNet using multiscopic images without post-processing. Details in the black boxes are zoomed.}
\label{fig:mccnn}
\vspace{-0.5cm}
\end{figure}

Since there is no available large dataset for multiscopic vision, we synthesize a dataset with hundreds of scenes using 3D render engine House3D \cite{wu2018building} and SceneNet RGB-D \cite{mccormac2017scenenet}. For each scene in our dataset, there are one center image and four surrounding images with resolution $1280\times1080$. Each color image has its corresponding ground-truth disparity map. The baseline varies from 0.05 to 0.2 in meters, and the maximum disparity is 255. Two example scenes are presented in Fig.~\ref{fig:synthetic}. For now, there are around 500 scenes of images, and we are keeping generating more data. This dataset can be utilized in many applications beyond multiscopic depth estimation: multiscopic super resolution, i.e. using multiscopic images to compute an image with much more resolutions; multiscopic view synthesis \cite{zhou2018stereo}, i.e. using a few of views to predict novel views like predicting the right view using left view and center view.

\begin{figure*}[t]
\centering
\begin{subfigure}{0.33\columnwidth}
  \centering
  \includegraphics[width=1.02\columnwidth, trim={0cm 0cm 0cm 0cm}, clip]{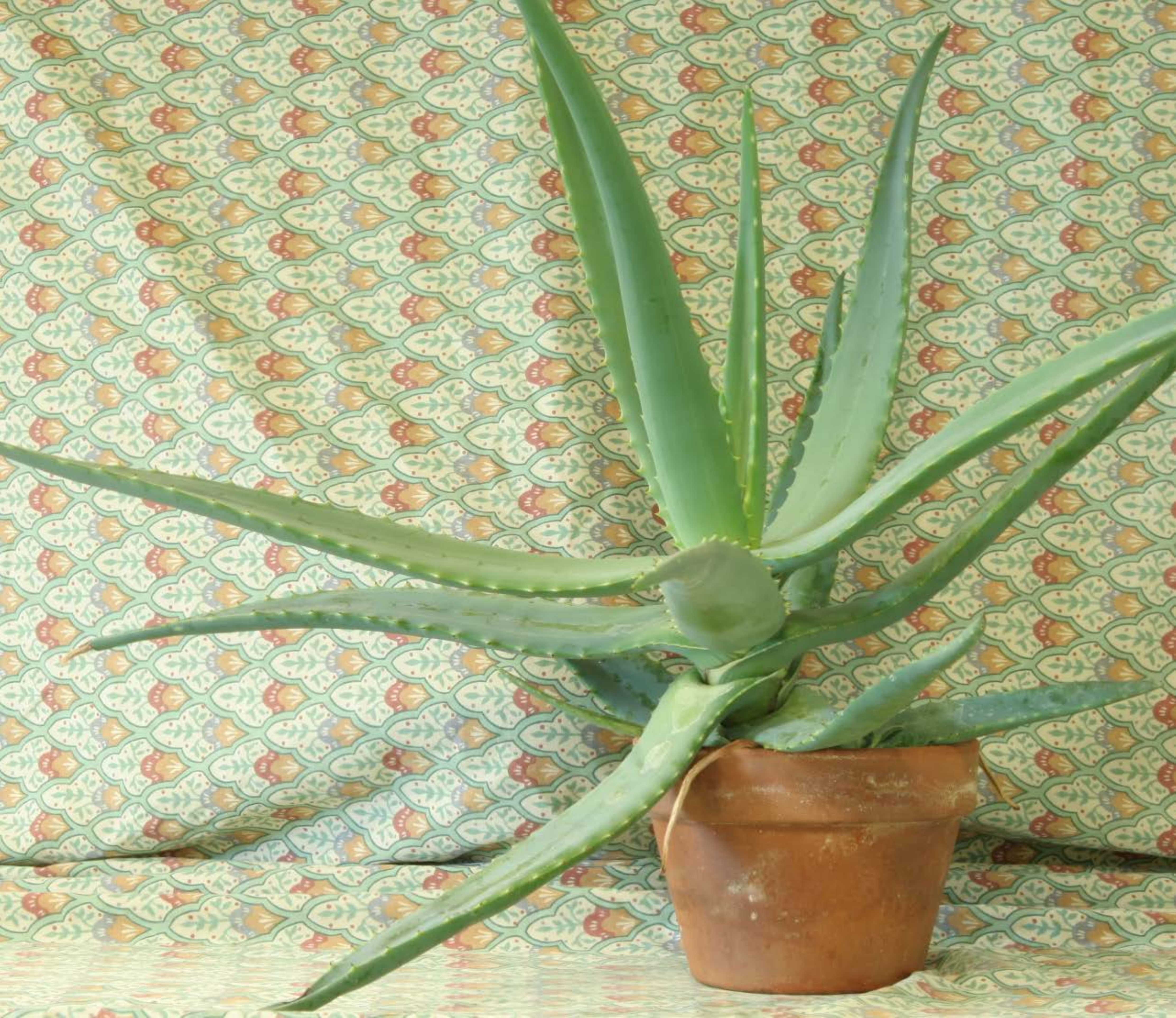}
%   \caption{Reference image}
\end{subfigure}
\begin{subfigure}{0.33\columnwidth}
  \centering
  \includegraphics[width=1.02\columnwidth, trim={0cm 0cm 0cm 0cm}, clip]{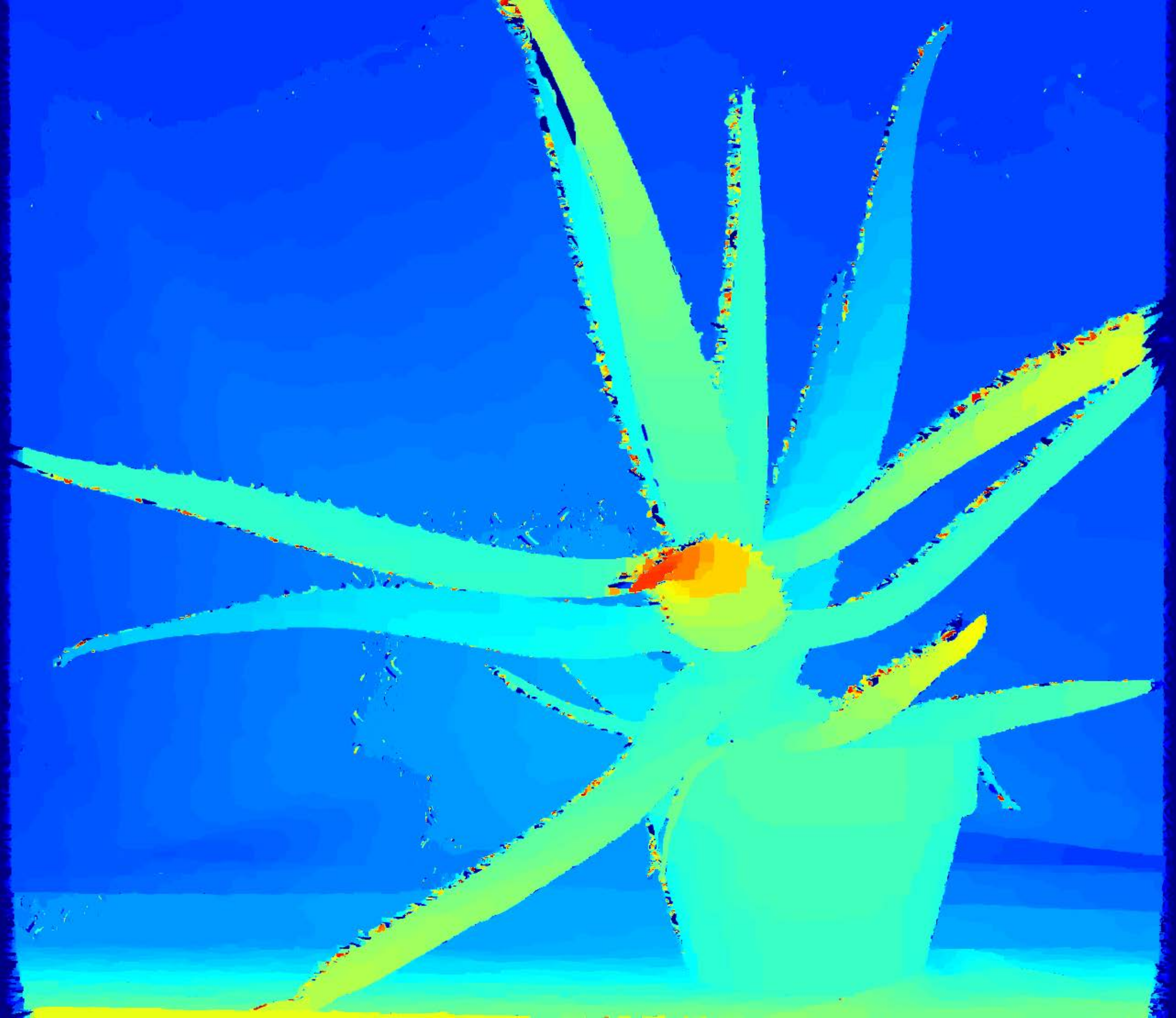}
%   \caption{Stereo GC}
\end{subfigure}
\begin{subfigure}{0.33\columnwidth}
  \centering
  \includegraphics[width=1.02\columnwidth, trim={0cm 0cm 0cm 0cm}, clip]{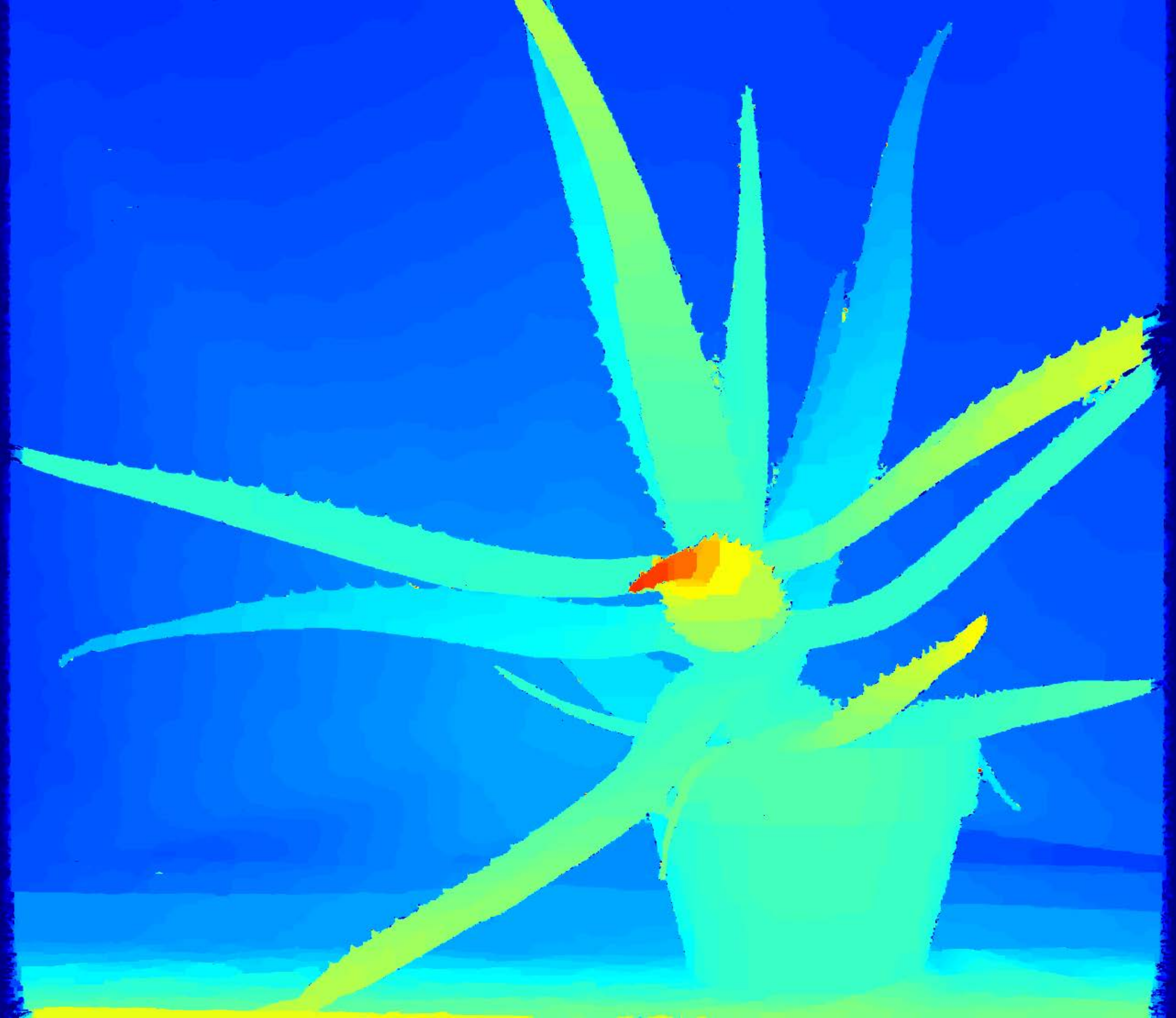}
%   \caption{Multiscopic GC}
\end{subfigure}
\begin{subfigure}{0.33\columnwidth}
  \centering
  \includegraphics[width=1.02\columnwidth, trim={0cm 0cm 0cm 0cm}, clip]{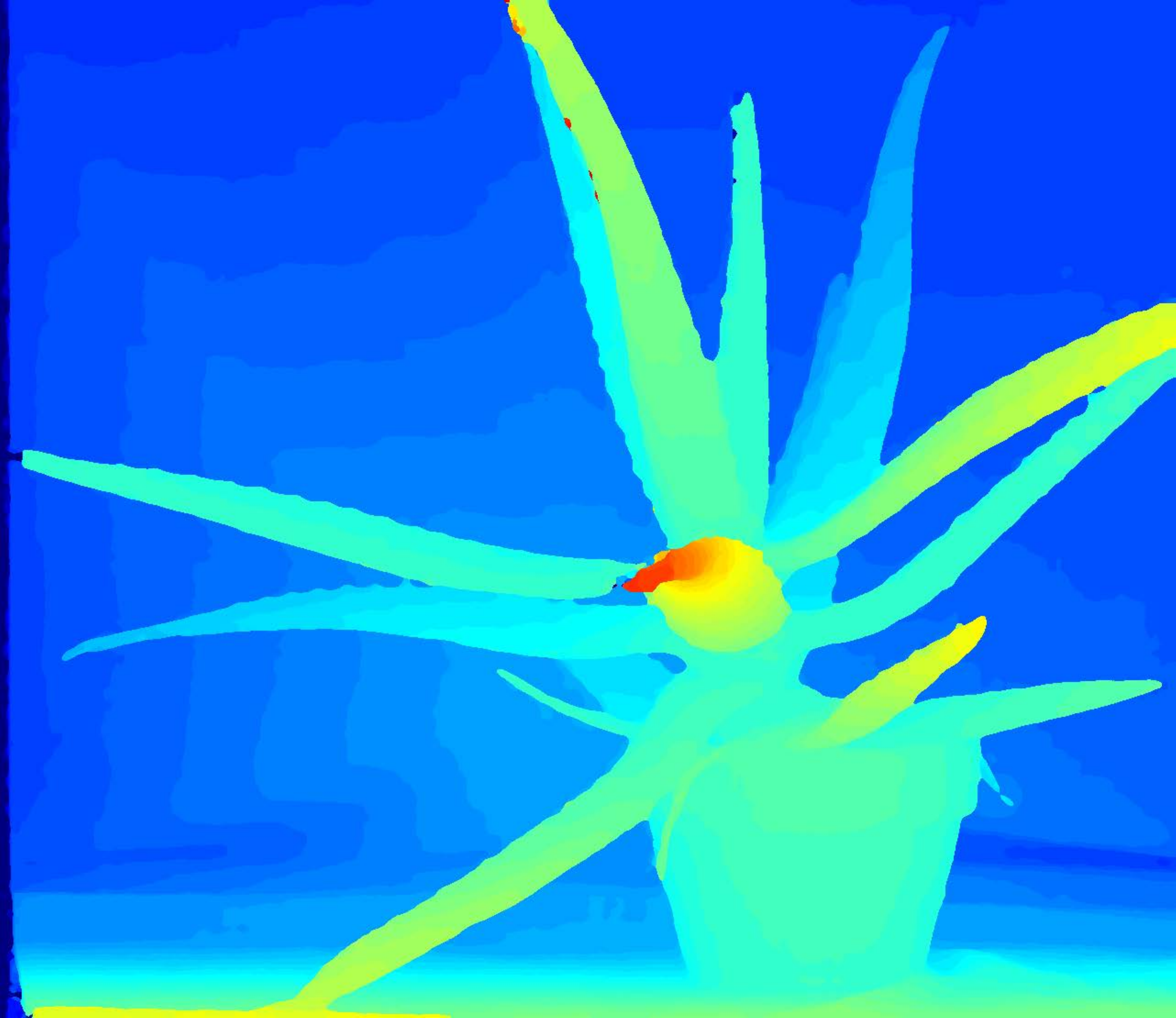}
%   \caption{Stereo MC-CNN}
\end{subfigure}
\begin{subfigure}{0.33\columnwidth}
  \centering
  \includegraphics[width=1.02\columnwidth, trim={0cm 0cm 0cm 0cm}, clip]{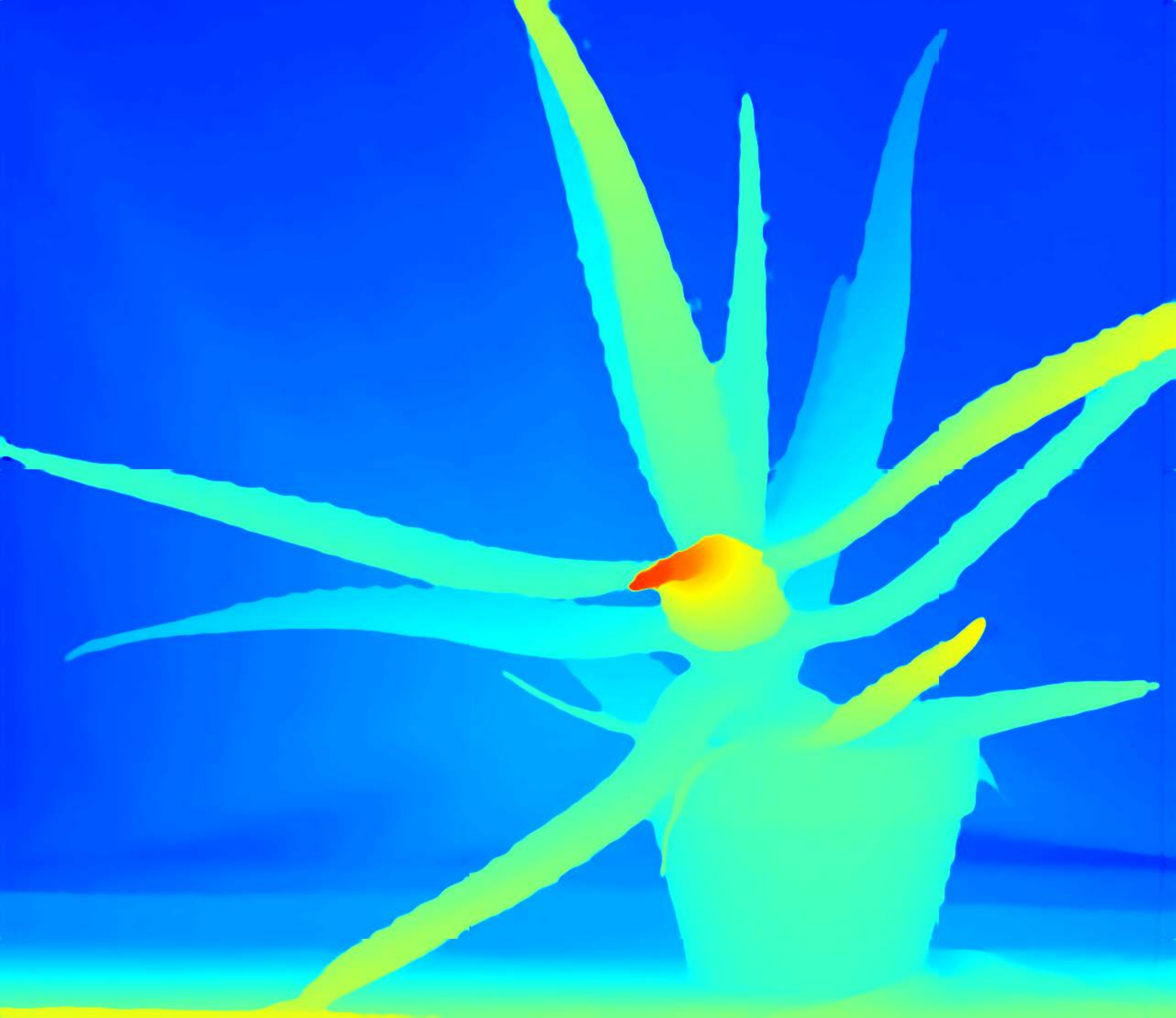}
%   \caption{MFuseNet Fusion}
\end{subfigure}
\begin{subfigure}{0.33\columnwidth}
  \centering
  \includegraphics[width=1.02\columnwidth, trim={0cm 0cm 0cm 0cm}, clip]{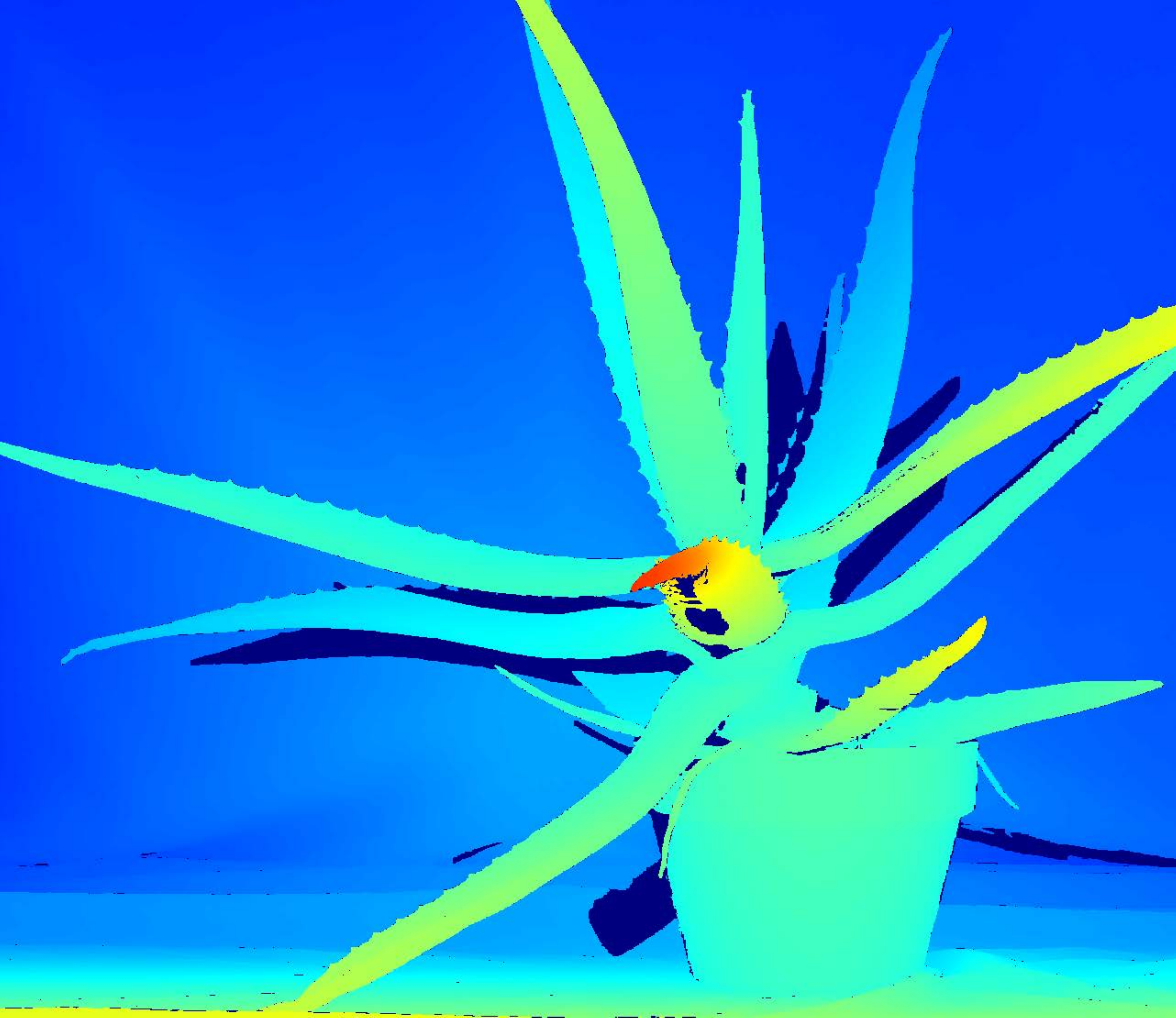}
%   \caption{Ground truth}
\end{subfigure}

\begin{subfigure}{0.33\columnwidth}
  \centering
  \includegraphics[width=1.02\columnwidth, trim={0cm 0cm 0cm 0cm}, clip]{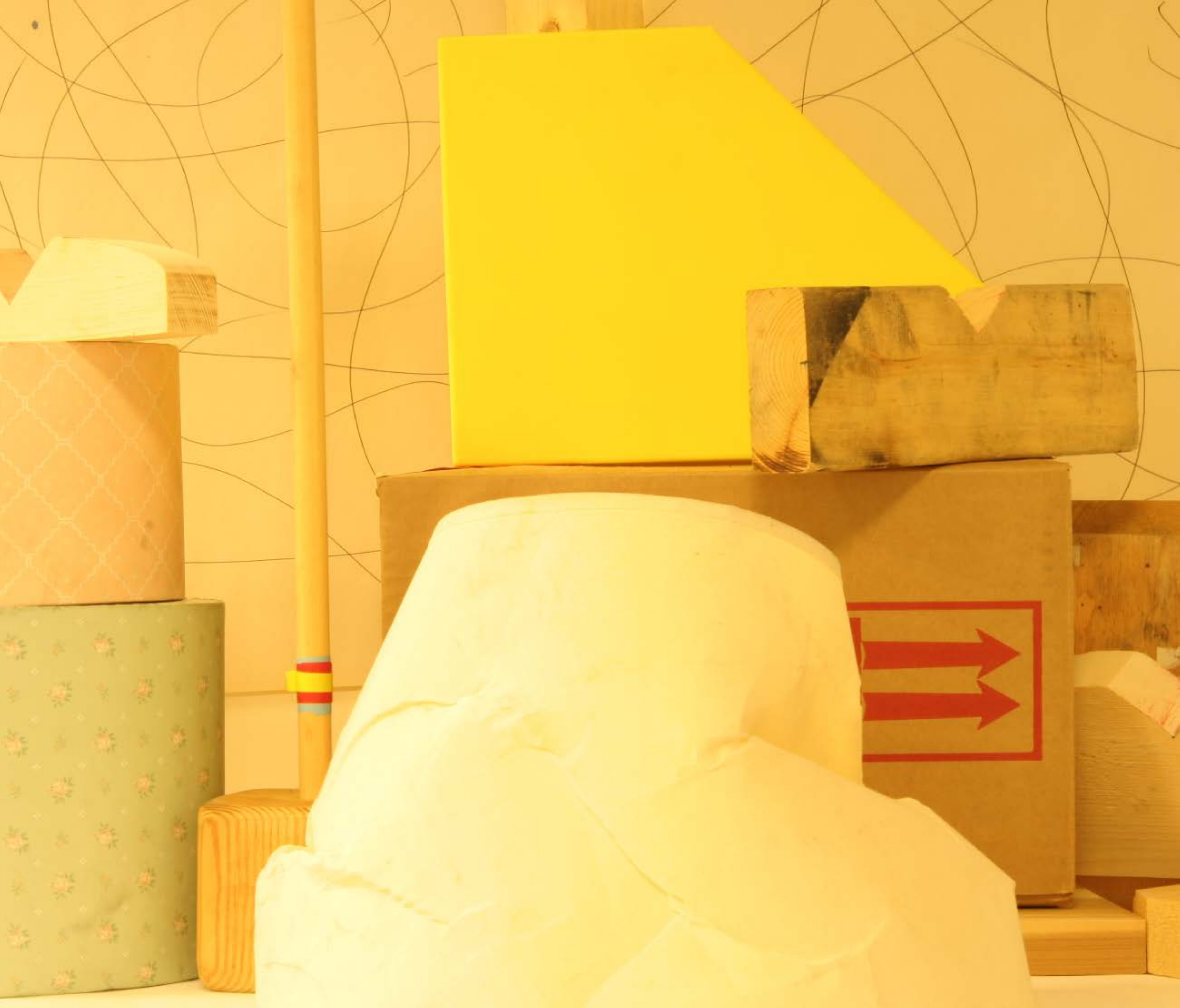}
  \caption{Reference image}
\end{subfigure}
\begin{subfigure}{0.33\columnwidth}
  \centering
  \includegraphics[width=1.02\columnwidth, trim={0cm 0cm 0cm 0cm}, clip]{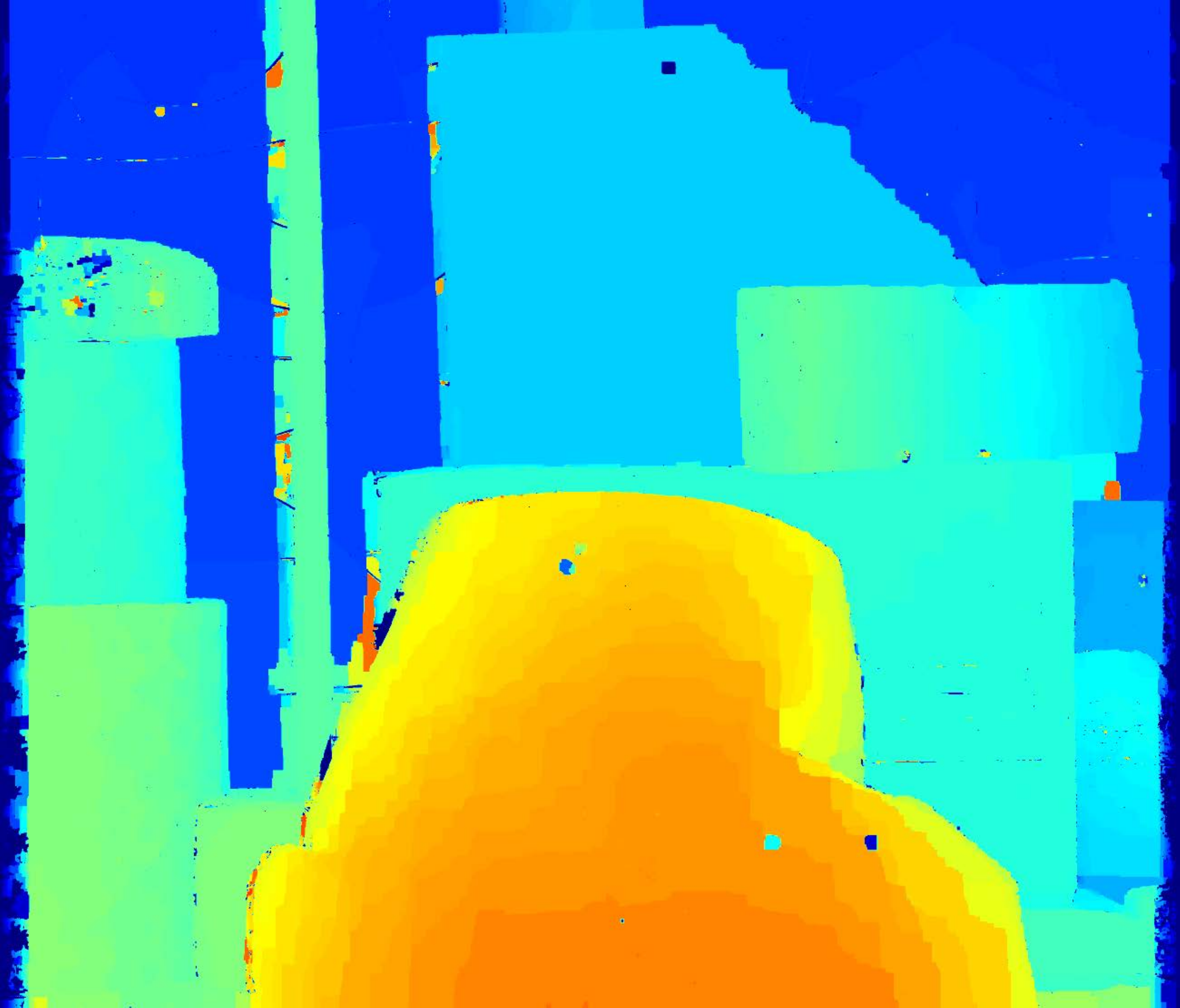}
  \caption{Stereo GC}
\end{subfigure}
\begin{subfigure}{0.33\columnwidth}
  \centering
  \includegraphics[width=1.02\columnwidth, trim={0cm 0cm 0cm 0cm}, clip]{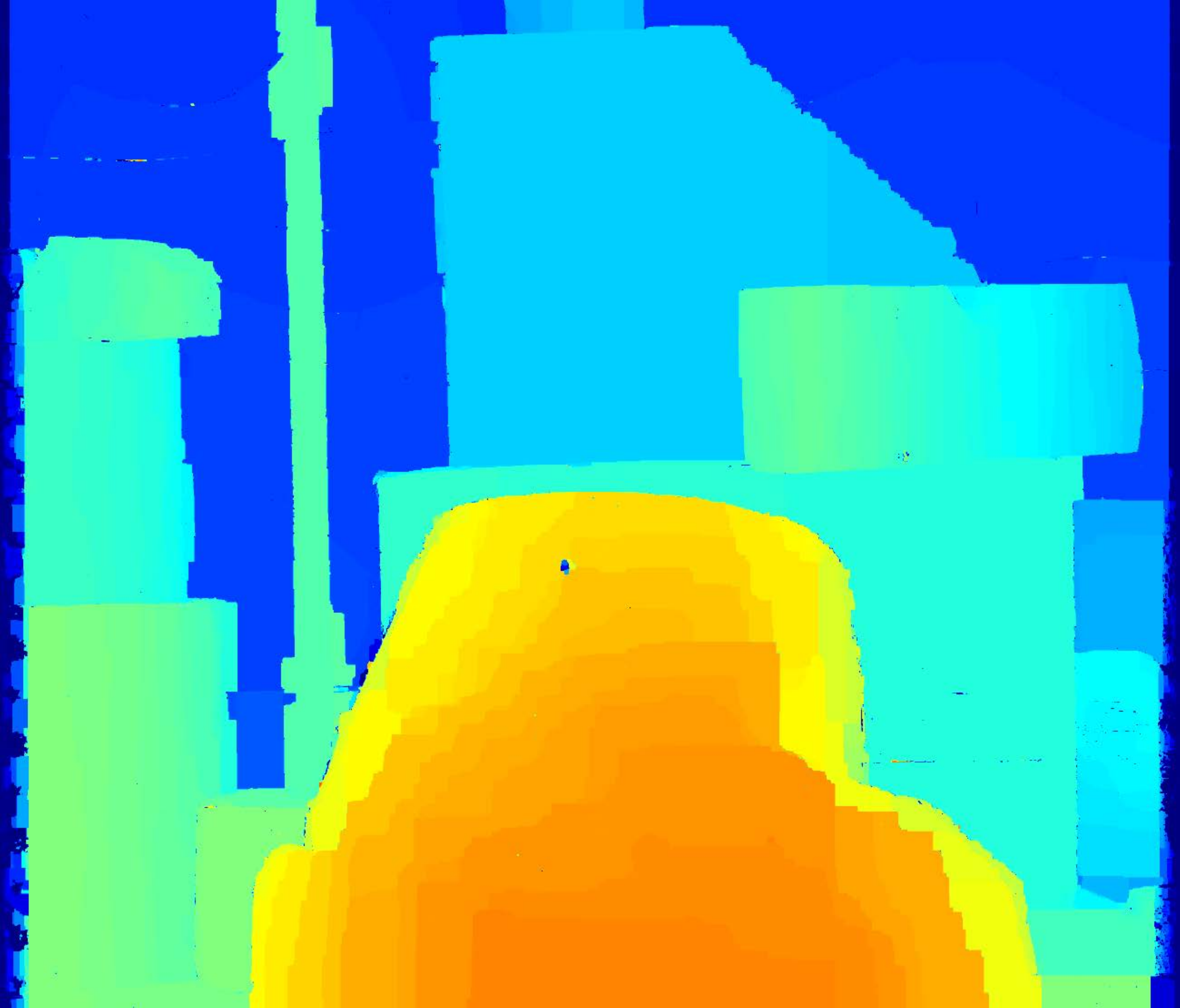}
  \caption{Multiscopic GC}
\end{subfigure}
\begin{subfigure}{0.33\columnwidth}
  \centering
  \includegraphics[width=1.02\columnwidth, trim={0cm 0cm 0cm 0cm}, clip]{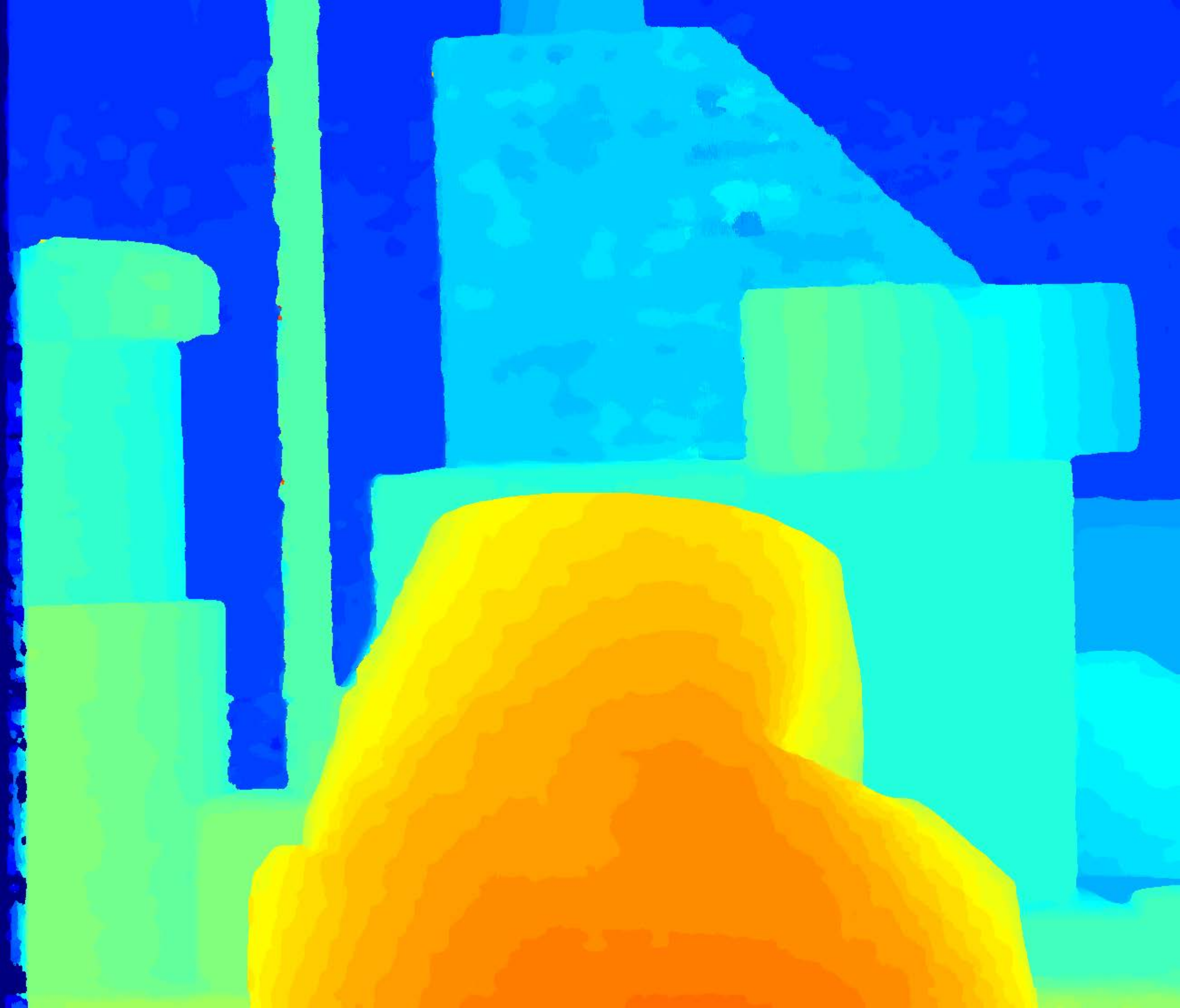}
  \caption{Stereo MC-CNN}
\end{subfigure}
\begin{subfigure}{0.33\columnwidth}
  \centering
  \includegraphics[width=1.02\columnwidth, trim={0cm 0cm 0cm 0cm}, clip]{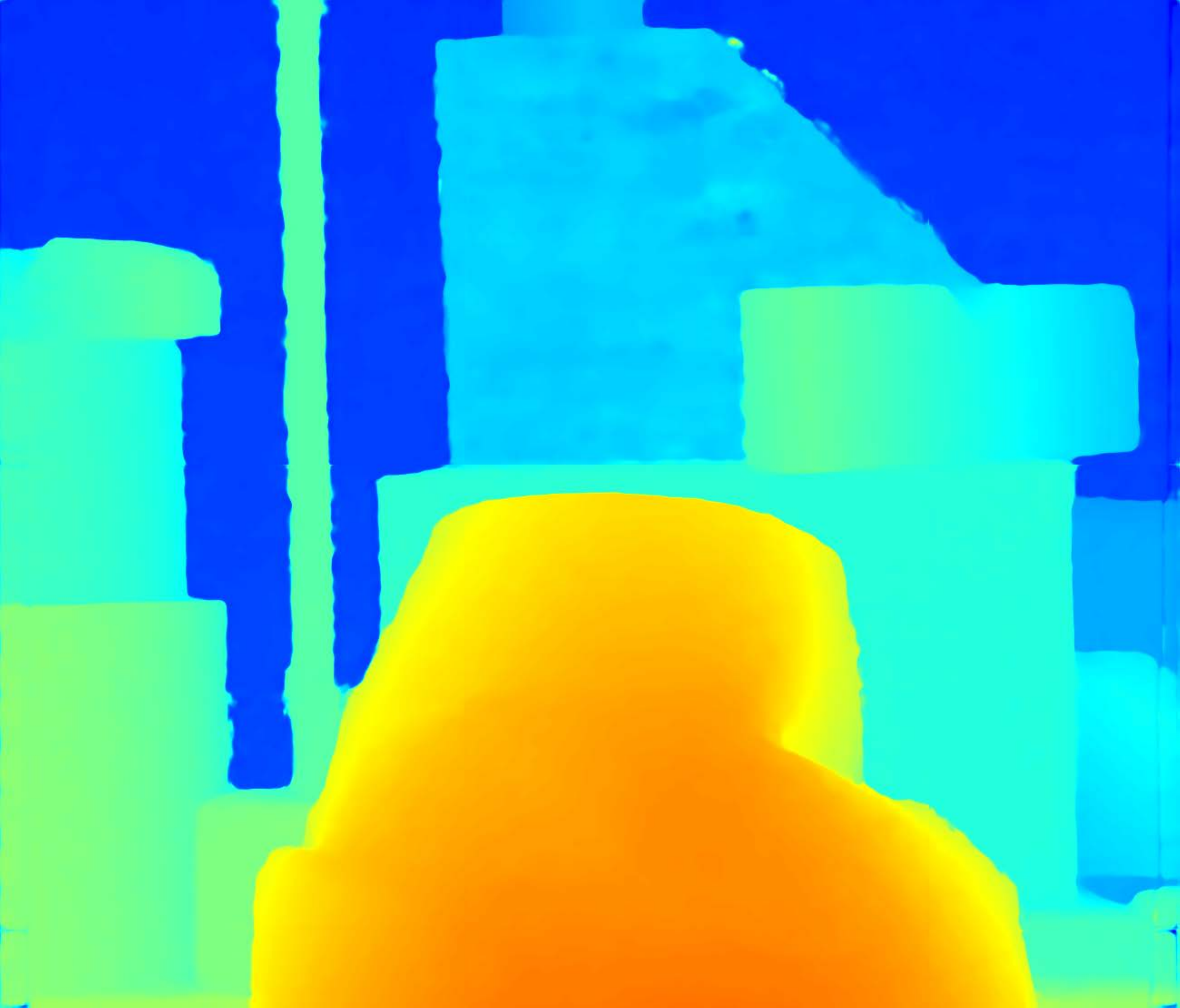}
  \caption{MFuseNet Fusion}
\end{subfigure}
\begin{subfigure}{0.33\columnwidth}
  \centering
  \includegraphics[width=1.02\columnwidth, trim={0cm 0cm 0cm 0cm}, clip]{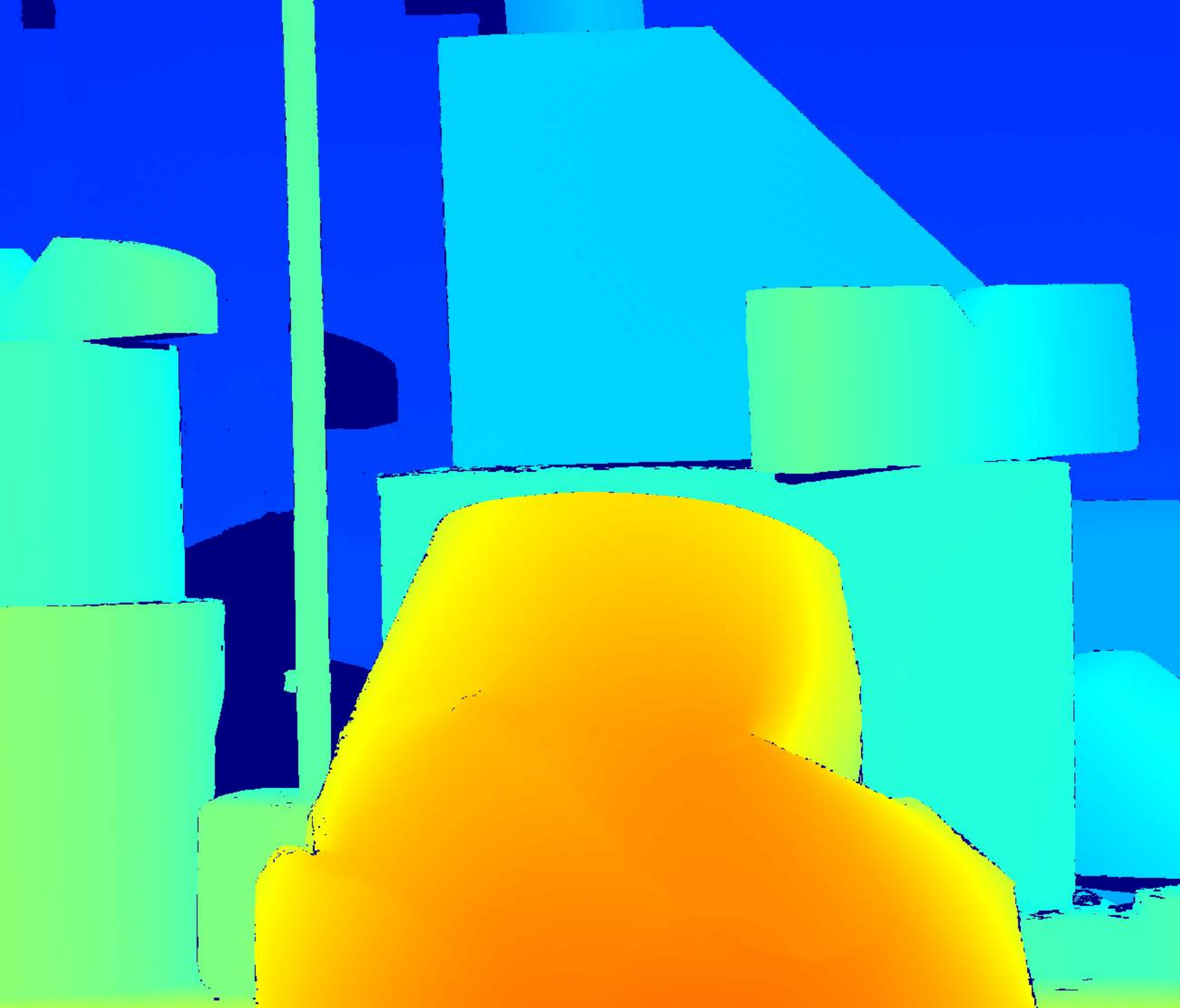}
  \caption{Ground truth}
\end{subfigure}

\caption{The disparity estimation results of different algorithms for two sets of images, Aloe and Lampshade in the Middlebury 2006 stereo dataset. The first image is the reference RGB image, i.e., the left image for stereo algorithms and the center image for multiscopic algorithms. Two images are used for stereo algorithms, and three images are used for multiscopic algorithms.}
\label{fig:exp_3frame}
\vspace{-0.2cm}
\end{figure*}

\begin{table*}[]
\centering
\begin{tabular}{r c c c c c}
\toprule
Methods & RMS & AvgErr & Bad$0.5$ & Bad$1$ & Bad$2$  \\
\midrule
Stereo GC & $3.946$ & $0.982$ & $10.31\%$ & $4.64\%$ & $3.55\%$\\
Heuristic Multiscopic GC & $2.195\ (\downarrow 44.4\%)$ & $0.539\ (\downarrow 45.1\%)$ & $8.49\%$ & $3.23\%$ & $1.79\%$\\
\midrule
Stereo MC-CNN & $3.760$ & $0.858$ & $9.35\%$ & $3.88\%$ & $3.16\%$\\
Multiscopic Heuristic Fusion & $1.485\ (\downarrow 60.5\%)$ & $0.424\ (\downarrow 50.58\%)$ & $5.79\%$ & $1.91\%$ & $1.28\%$\\
Stereo MFuseNet & $1.514\ (\downarrow 59.73\%)$ & $0.365\ (\downarrow 57.46\%)$ & $7.11\%$ & $3.30\%$ & $2.08\%$\\
Multiscopic MFuseNet Fusion & $\mathbf{1.116\ (\downarrow 70.32\%)}$ & $\mathbf{0.262\ (\downarrow 69.46\%)}$ & $\mathbf{3.29\%}$ & $\mathbf{1.89\%}$ & $\mathbf{1.20\%}$\\
\bottomrule
\end{tabular}
\caption{Matching results on 21 scenes in the Middlebury 2006 stereo dataset \cite{hirschmuller2007evaluation}. }
\label{tab:results}
\vspace{-0.3cm}
\end{table*}

\subsection{Network Training}

To train the lightweight MFuseNet with the synthetic dataset, we only use 27 scenes with maximum disparity 60 (maximum disparity in the Middlebury dataset). Although the number of scenes is relatively small, the network can be well optimized and performs stably on unseen scenarios. 
Two disparity maps obtained by MFuseNet and the corresponding synthetic images used during training are shown in Fig.~\ref{fig:synthetic}. The real-world result on the multiscopic images captured by our monocular perception system is presented in Fig.~\ref{fig:mccnn}. Compared with the output of stereo MC-CNN, the disparity map by MFuseNet is cleaner and smoother. In addition, the object boundary appears sharper in our disparity result.

% !TEX root =  ../main.tex

\section{EXPERIMENTS}
\label{sec:exp}

In this section, we present the details of our system setup and experiments results. The quantitative evaluation on the Middlebury 2006 stereo dataset and the qualitative experiments on real robots are demonstrated.

\subsection{System Setup}

\begin{figure}[]
\centering
  \includegraphics[width=0.6\columnwidth, trim={0cm 0cm 0cm 0cm}, clip]{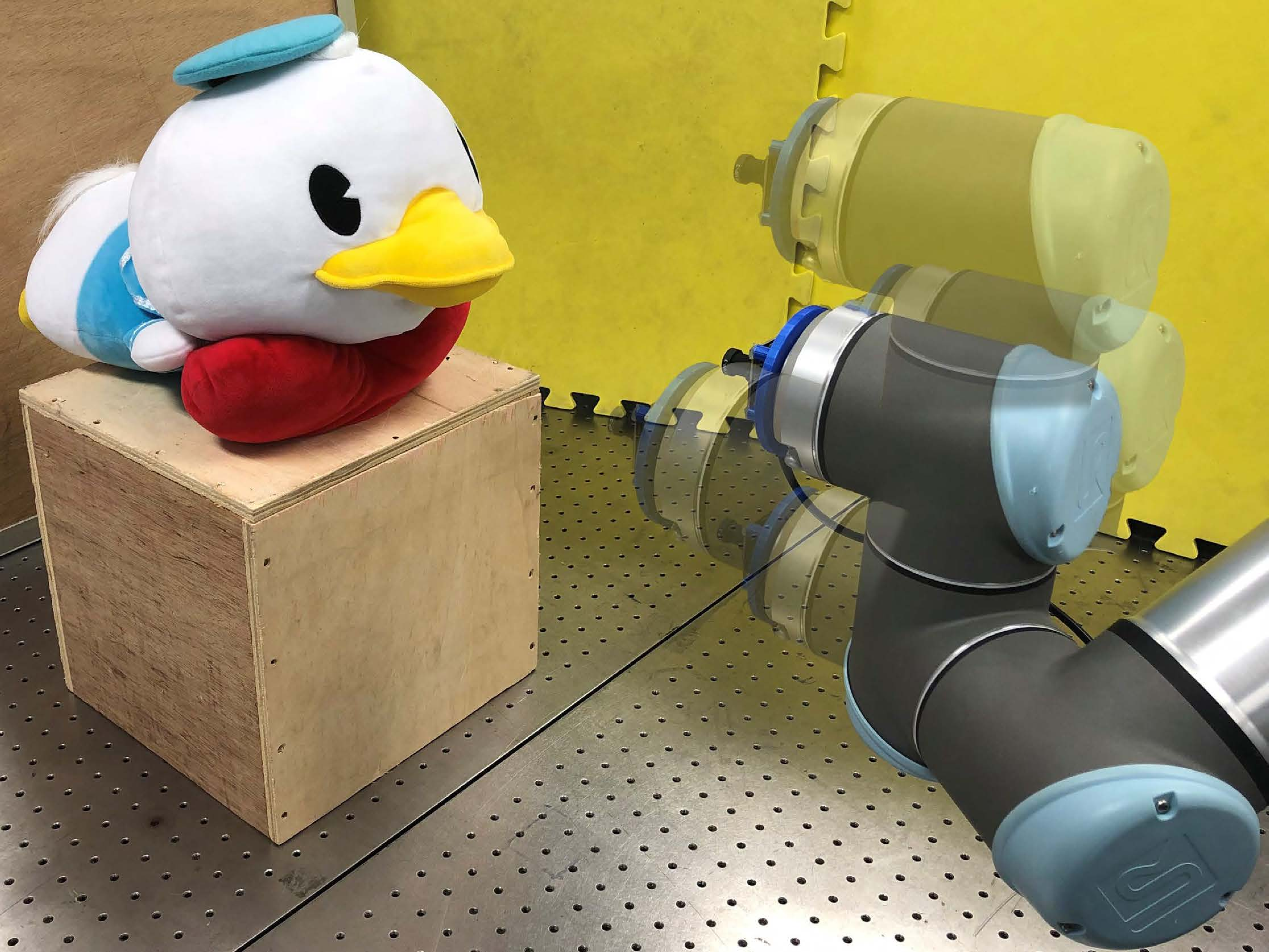}
\caption{The camera is mounted at the end of a robot arm and moved horizontally and vertically to take pictures from different views.}
\label{fig:robot}
\vspace{-0.5cm}
\end{figure}

To build the multiscopic vision system, we mount a monocular camera at the end of a robot arm, as displayed in Fig.~\ref{fig:robot}. The sensor we use is an ordinary USB video camera with Sony IMX322 inside, whose resolution is $1920\times1080$. The robot arm is UR10, a collaborative industrial robot whose repeatability is $\pm0.1\ \text{mm}$. UR10 has six rotating joints, so the end has 6 degrees of freedom. Hence the camera can move freely with any pose.

To capture a series of images with multiscopic structure, we command the UR10 to move the camera in its image plane, generating a set of co-planar images. For every movement with the same distance, we take one picture of the environment. Our system can only handle static scenes because it takes time to capture multiple images with a robot arm. We can take multiple images, and each of these images has the same parallax with its adjacent images. For example, we can take 9 images in  3 rows and 3 columns, which forms a multiscopic array. Also, we can adjust the baseline according to the need. For the sake of simplicity, we use five images in the real robot experiments to estimate disparity.

Our algorithm is run on a computer with an Nvidia GPU of GeForce GTX 1080 Ti. To evaluate the performance of our multiscopic algorithm, we conduct a quantitative evaluation on the Middlebury 2006 stereo dataset \cite{hirschmuller2007evaluation}, which contains calibrated and rectified image sequence for depth estimation. For both heuristic fusion and MFuseNet fusion, we use only three adjacent images to compute the disparity. Note that we can use our system to capture more images and do multiscopic matching with five or even more images.

% \begin{figure}[tbh]
% \centering
% \begin{subfigure}{0.8\columnwidth}
%   \centering
%   \includegraphics[width=1\columnwidth, trim={0cm 0cm 0cm 0cm}, clip]{figures/zoomin_noise.pdf}
%   \vspace{-0.5cm}
%   \caption{Noise}
% \end{subfigure}
% \begin{subfigure}{0.8\columnwidth}
%   \centering
%   \includegraphics[width=1\columnwidth, trim={0cm 0cm 0cm 0cm}, clip]{figures/zoomin_occlu.pdf}
%   \vspace{-0.5cm}
%   \caption{Occlusion}
% \end{subfigure}
% \begin{subfigure}{0.8\columnwidth}
%   \centering
%   \includegraphics[width=1\columnwidth, trim={0cm 0cm 0cm 0cm}, clip]{figures/zoomin_reflec.pdf}
%   \vspace{-0.5cm}
%   \caption{Reflection}
% \end{subfigure}
% \caption{The visual comparison between stereo matching graph cuts (I) and multiscopic graph cuts (II) in noisy, occluded and reflective areas.}
% \label{fig:zoom}
% \end{figure}

\begin{figure*}[]
\centering
\begin{subfigure}{0.33\columnwidth}
  \centering
  \includegraphics[width=1\columnwidth, trim={25cm 5cm 24cm 14cm}, clip]{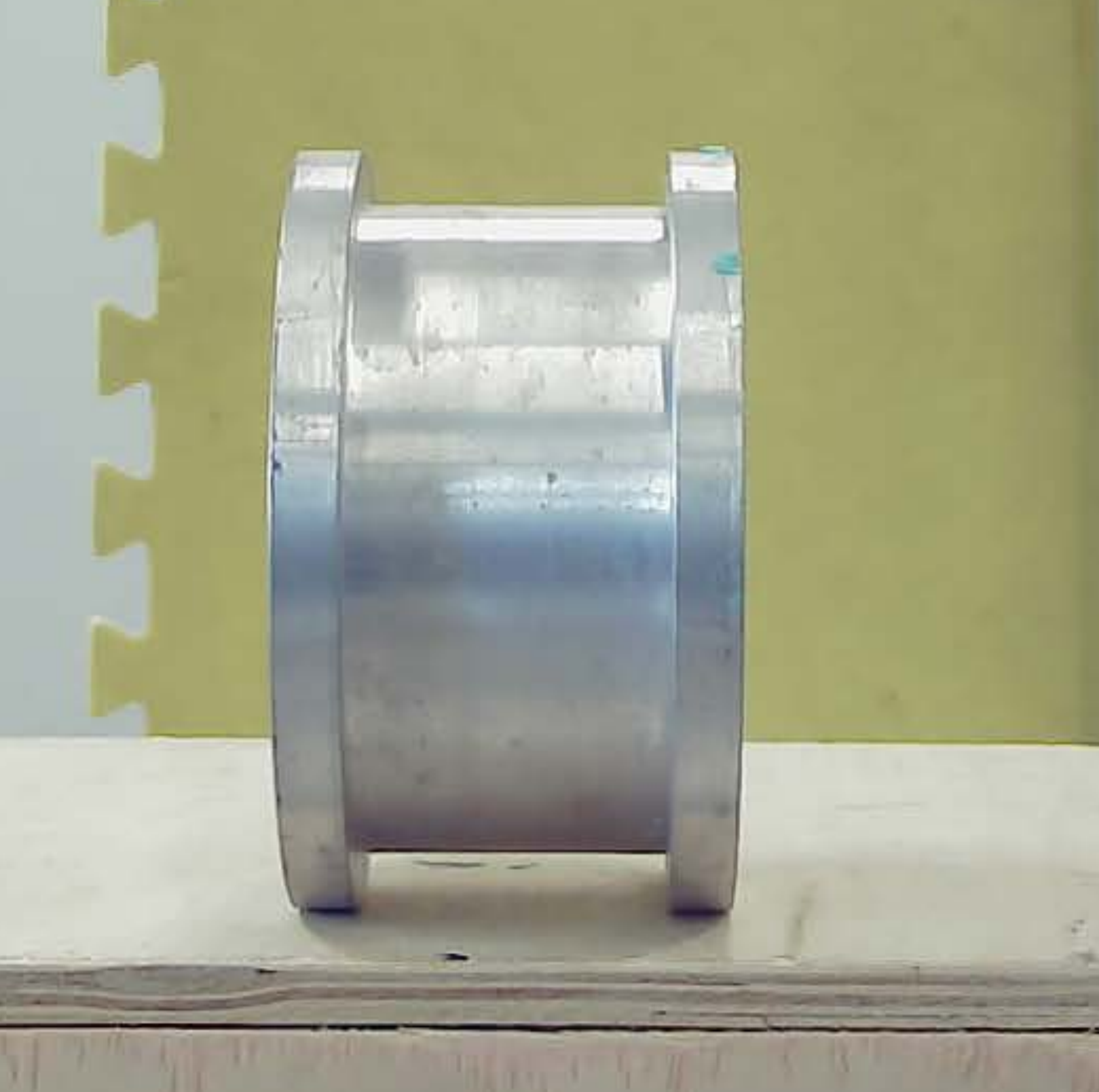}
  \caption{Center image}
\end{subfigure}
\centering
\begin{subfigure}{0.33\columnwidth}
  \centering
  \includegraphics[width=1\columnwidth, trim={25cm 5cm 24cm 14cm}, clip]{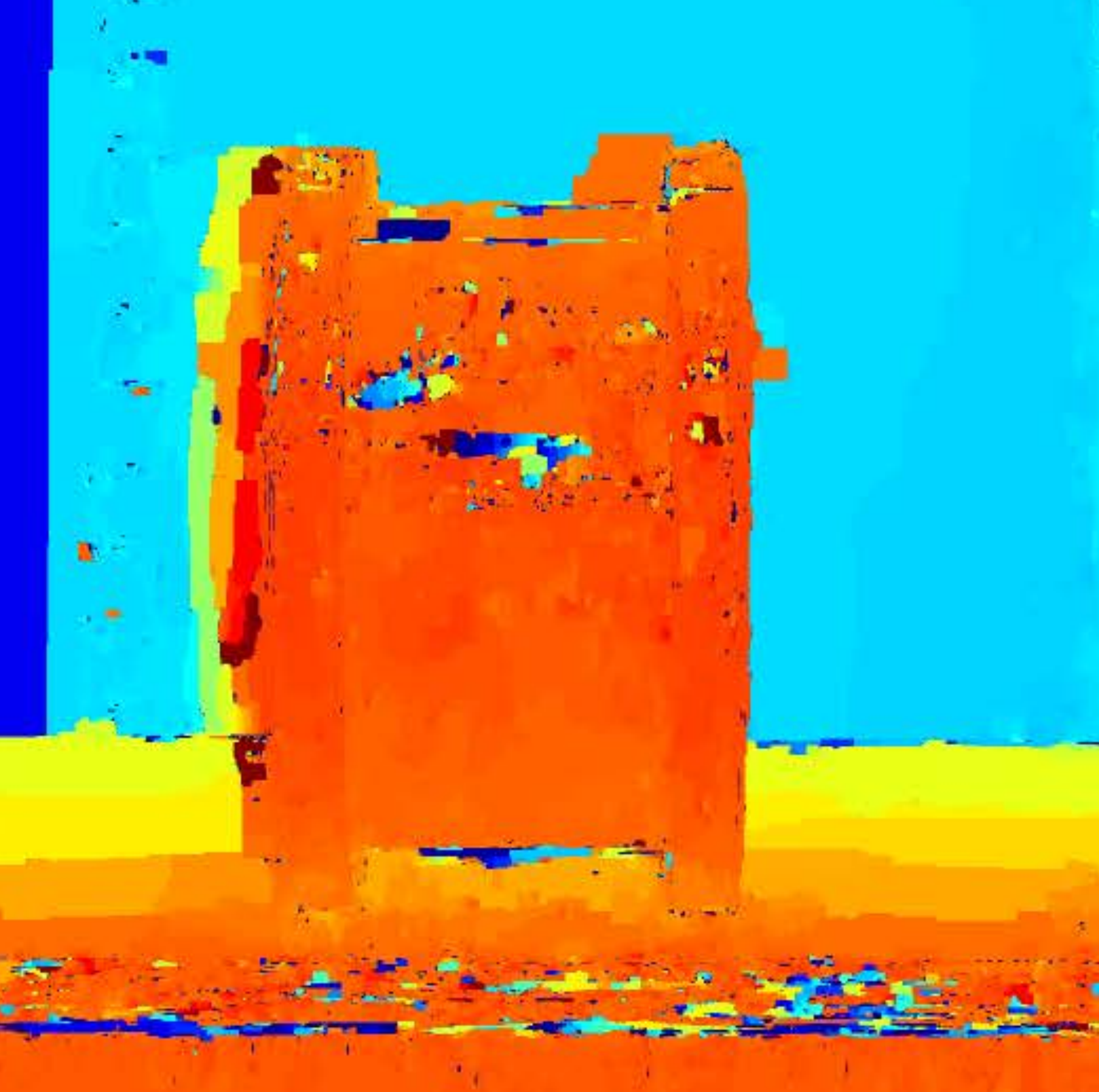}
  \caption{Stereo GC}
\end{subfigure}
\begin{subfigure}{0.33\columnwidth}
  \centering
  \includegraphics[width=1\columnwidth, trim={25cm 5cm 24cm 14cm}, clip]{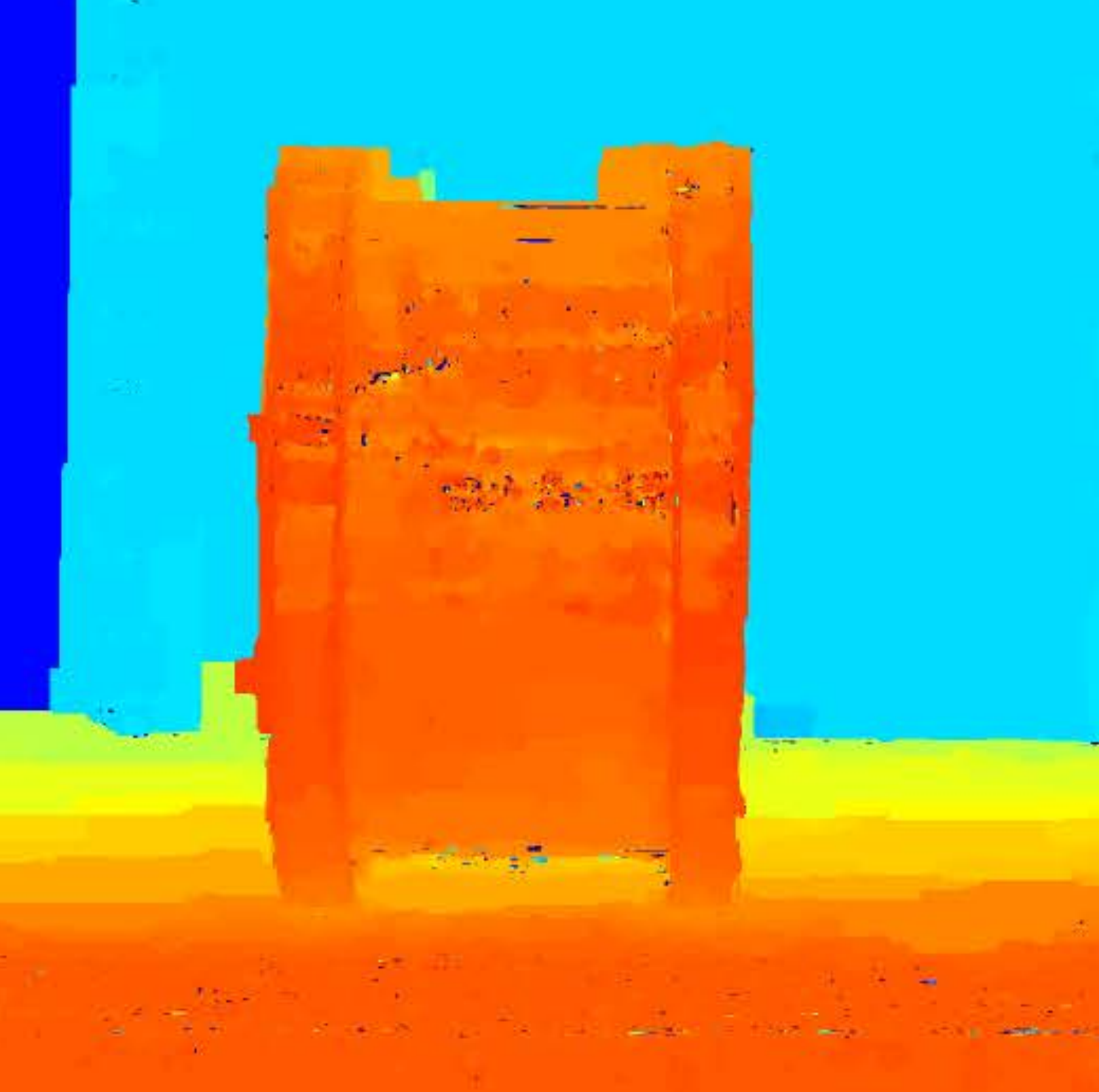}
  \caption{Multiscopic GC}
\end{subfigure}
\begin{subfigure}{0.33\columnwidth}
  \centering
  \includegraphics[width=1\columnwidth, trim={11cm 5cm 10cm 7cm}, clip]{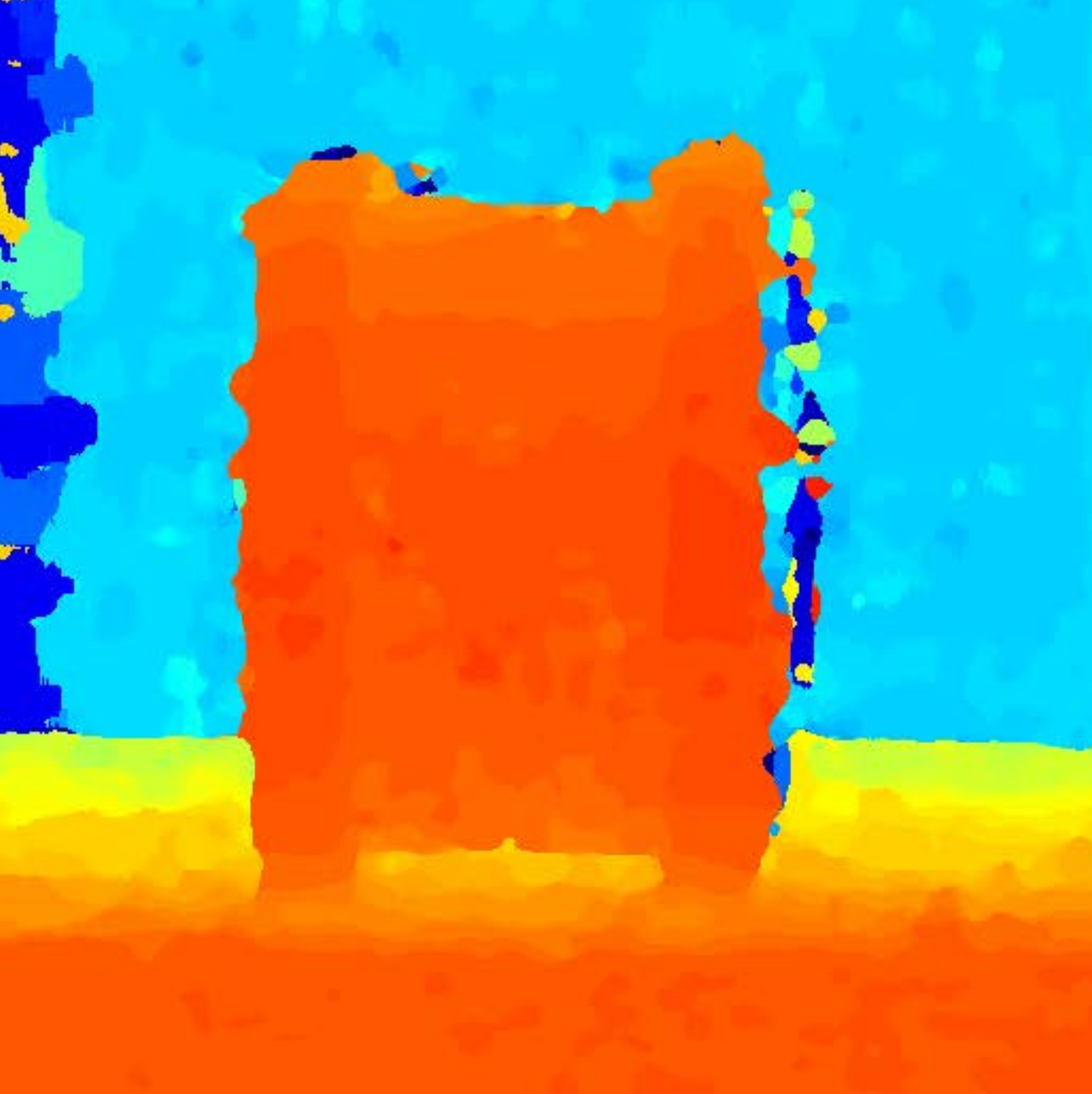}
  \caption{Stereo MC-CNN}
\end{subfigure}
\begin{subfigure}{0.33\columnwidth}
  \centering
  \includegraphics[width=1\columnwidth, trim={11cm 5cm 10cm 7cm}, clip]{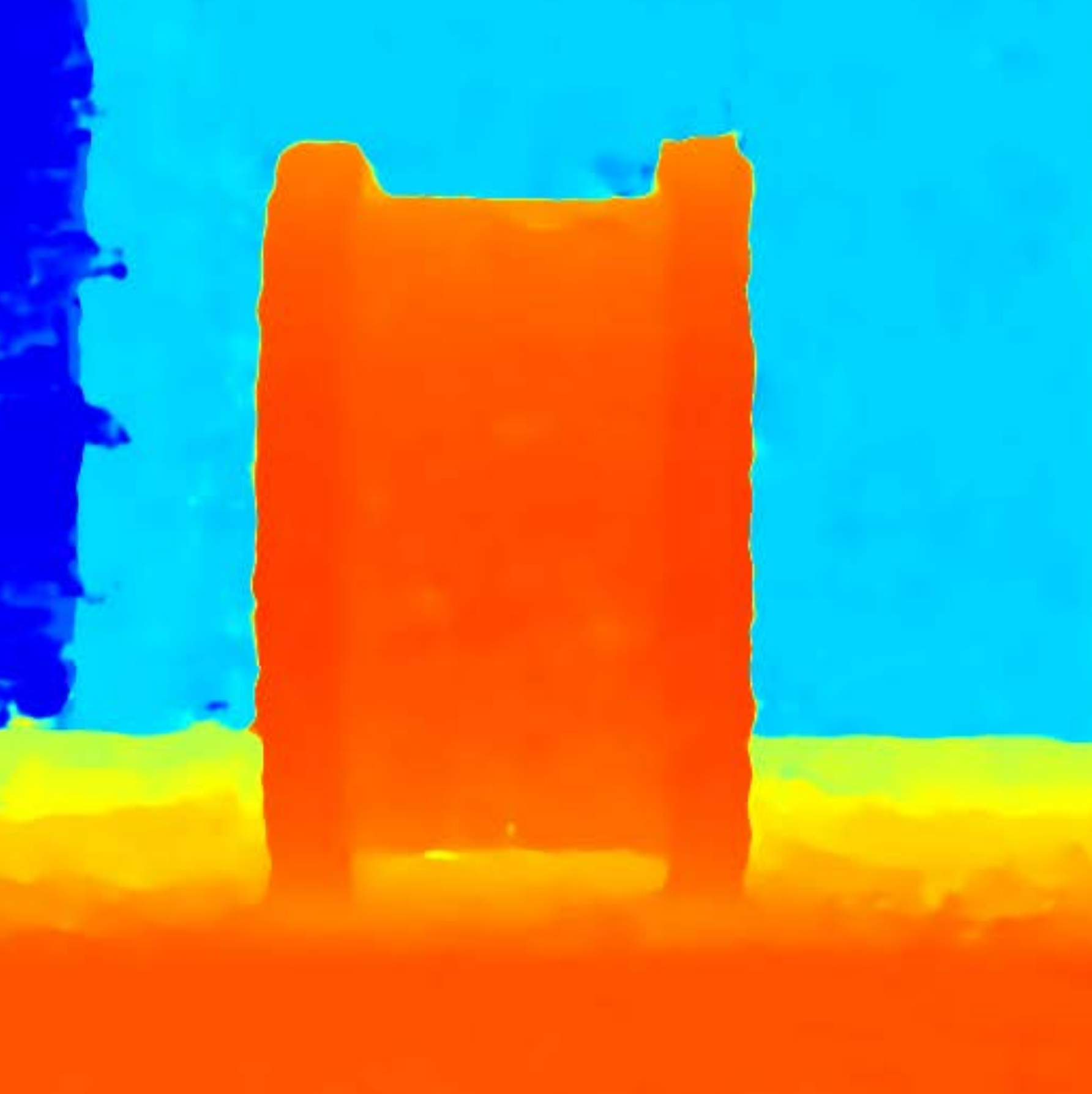}
  \caption{MFuseNet}
\end{subfigure}
\caption{The disparity estimation results of different algorithms for a reflective workpiece.}
\label{fig:exp_5frame}
\vspace{-0.2cm}
\end{figure*}

\subsection{Evaluation on Middleburry}

%The images in the Middlebury Stereo Datasets are well calibrated and rectified, so it can better show the improvement of multiscopic matching without the influence of system error. But since there is only image sequence in horizontal in this dataset, we choose only three images, the view 0, view 1, view 2 as the left image, center image and right image. The baseline between view 1 and view 0 or view 2 is 40 mm. Thus with the center image as reference, there will be two costs, one is between the left image and the center image, the other is between the right image and the center image. Because these images are rectified, there is no much noise for the smaller one of these two costs. Thus the fusion of these two costs can be directly using the smaller one for both block matching and graph cuts matching as Equ.~\ref{equ:min}.

The images in the Middlebury dataset are well calibrated and rectified, so it can quantitatively show the improvement of multiscopic matching without the influence of image calibration error. Since there are only images captured in the horizontal direction in this dataset, we choose the first three images. We use the first image as the left image, the second one as the center image, and the third image as the right image for the multiscopic algorithm. The baseline between two adjacent images is 40 mm. With the center image as a reference image, there are two cost volumes to be combined. One is between the left image and the center image, and the other one is between the right image and the center image. Because there are only two cost volumes, the heuristic fusion can be directly using the smaller one according to Equ.~\ref{equ:min}.

The maximum searching disparity is set to $60$ and the minimum is set to $1$. For graph cuts, the occlusion penalty $K$ is set to $10$ and the smoothness parameters $\lambda_1, \lambda_2, \theta, d_{\text{CUTOFF}}$ are set to $9, 3, 8, 5$, respectively. 
For MC-CNN, the pre-trained accurate Middlebury network model is used in the experiment. Then based on the cost volumes obtained from MC-CNN,  MFuseNet is trained using the left, center, and right images from 27 synthetic scenes with maximum disparity 60 in the multiscopic mode, and using the center and right images in the stereo mode. All the results are obtained without any post-processing.

We use five metrics to evaluate the matching results, as is summarized in TABLE \ref{tab:results}. The RMS is the root-mean-square error, AvgErr is the average absolute error, Bad0.5 is the percentage of pixels whose error is greater than 0.5. Bad1 and Bad2 are defined similarly. It can be seen from these five metrics that the multiscopic framework can improve the correspondence matching a lot even with only three images. Also, MFuseNet fusion can outperform heuristic fusion significantly. The average decrease of the average absolute error on 21 Middlebury scenes can reach $69.5\%$, and the one of root-mean-square error is $70.3\%$. Compared with heuristic fusion, MFuseNet fusion can achieve around another $40\%$ absolute error decrease. Even though MFuseNet is trained with only 27 sets of synthetic images, it can generalize and perform well in different scenes. 

For qualitative comparison, we randomly choose two scenes of images from the Middlebury dataset, Aloe and Lampshade.  In Fig.~\ref{fig:exp_3frame}, we show the disparity results of stereo graph cuts, multiscopic heuristic graph cuts, stereo MC-CNN, and multiscopic MFuseNet fusion without any post-processing. Compared to the stereo matching, multiscopic matching produces less noise and better reconstruction in occluded areas. Compared to heuristic fusion, MFuseNet produces cleaner and smoother estimation, although the heuristic graph cuts generates sharper edges of some objects.

\subsection{Evaluation on Tsukuba Multiview Data}

\begin{figure}[]
\centering
\begin{subfigure}{0.45\columnwidth}
  \includegraphics[width=1\columnwidth, trim={0cm 0cm 0cm 0cm}, clip]{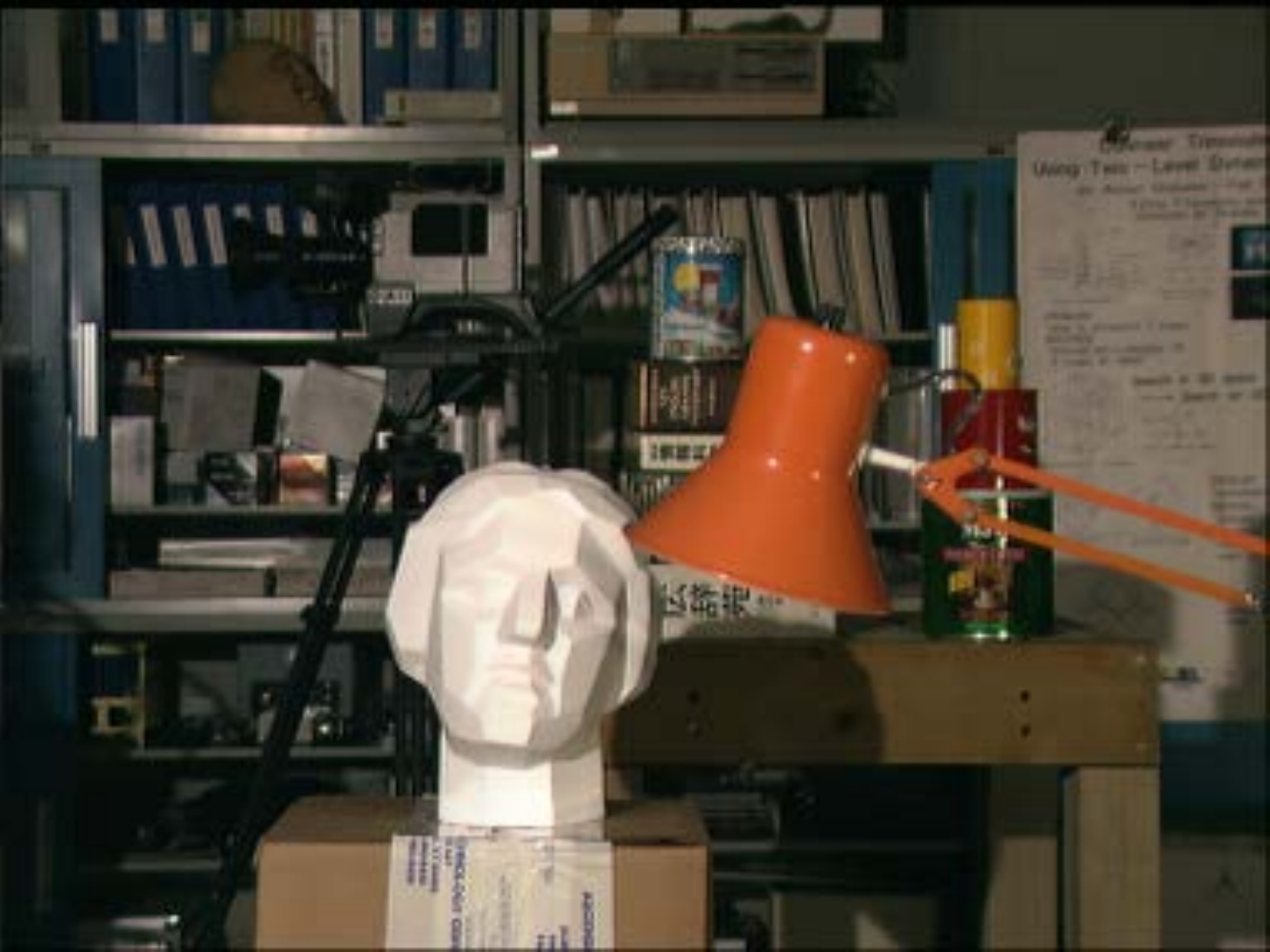}
  \caption{Tsukuba}
\end{subfigure}
\begin{subfigure}{0.45\columnwidth}
  \includegraphics[width=1\columnwidth, trim={0cm 0cm 0cm 0cm}, clip]{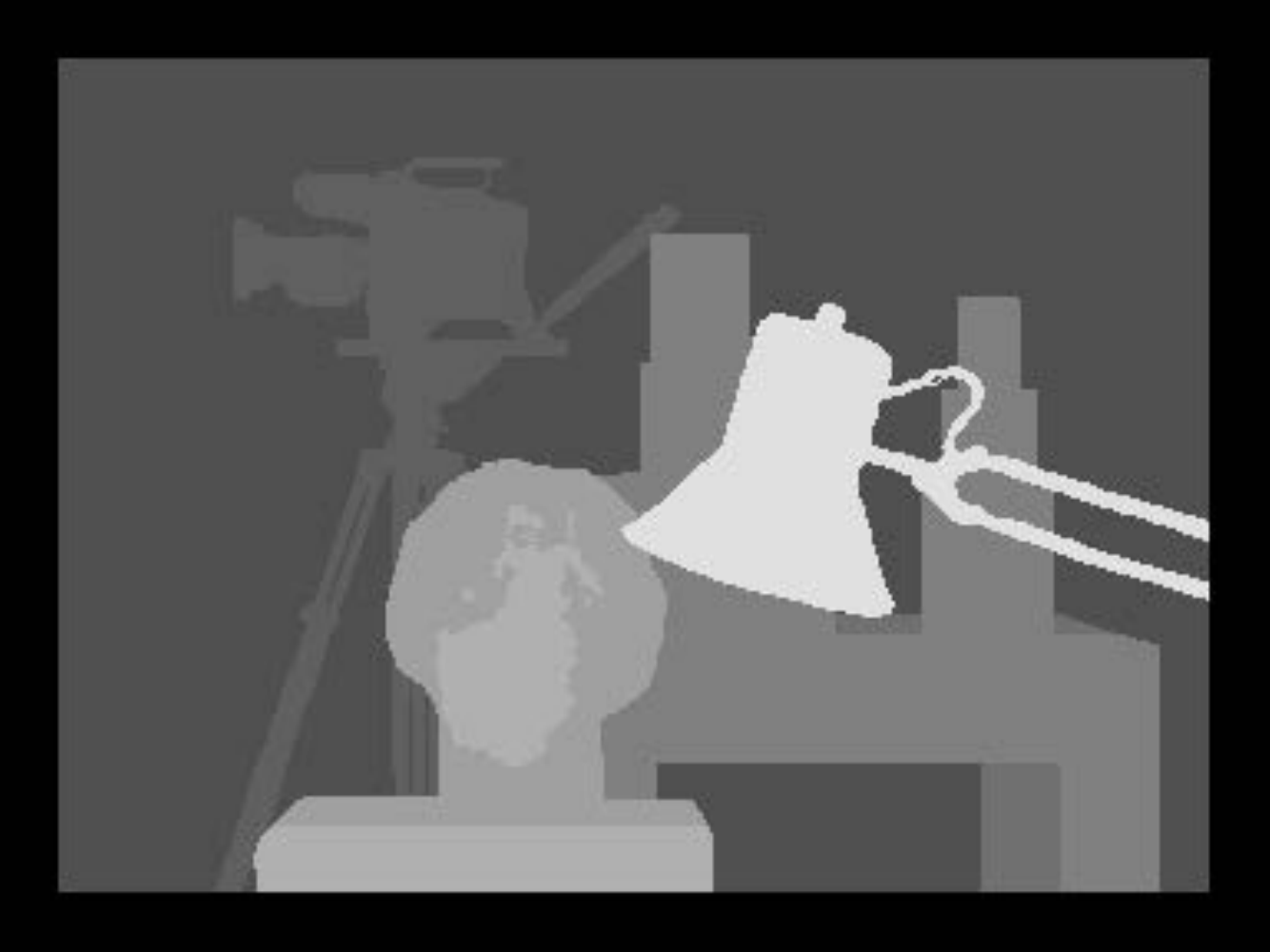}
  \caption{Ground truth}
\end{subfigure}
\begin{subfigure}{0.45\columnwidth}
  \includegraphics[width=1\columnwidth, trim={0cm 0cm 0cm 0cm}, clip]{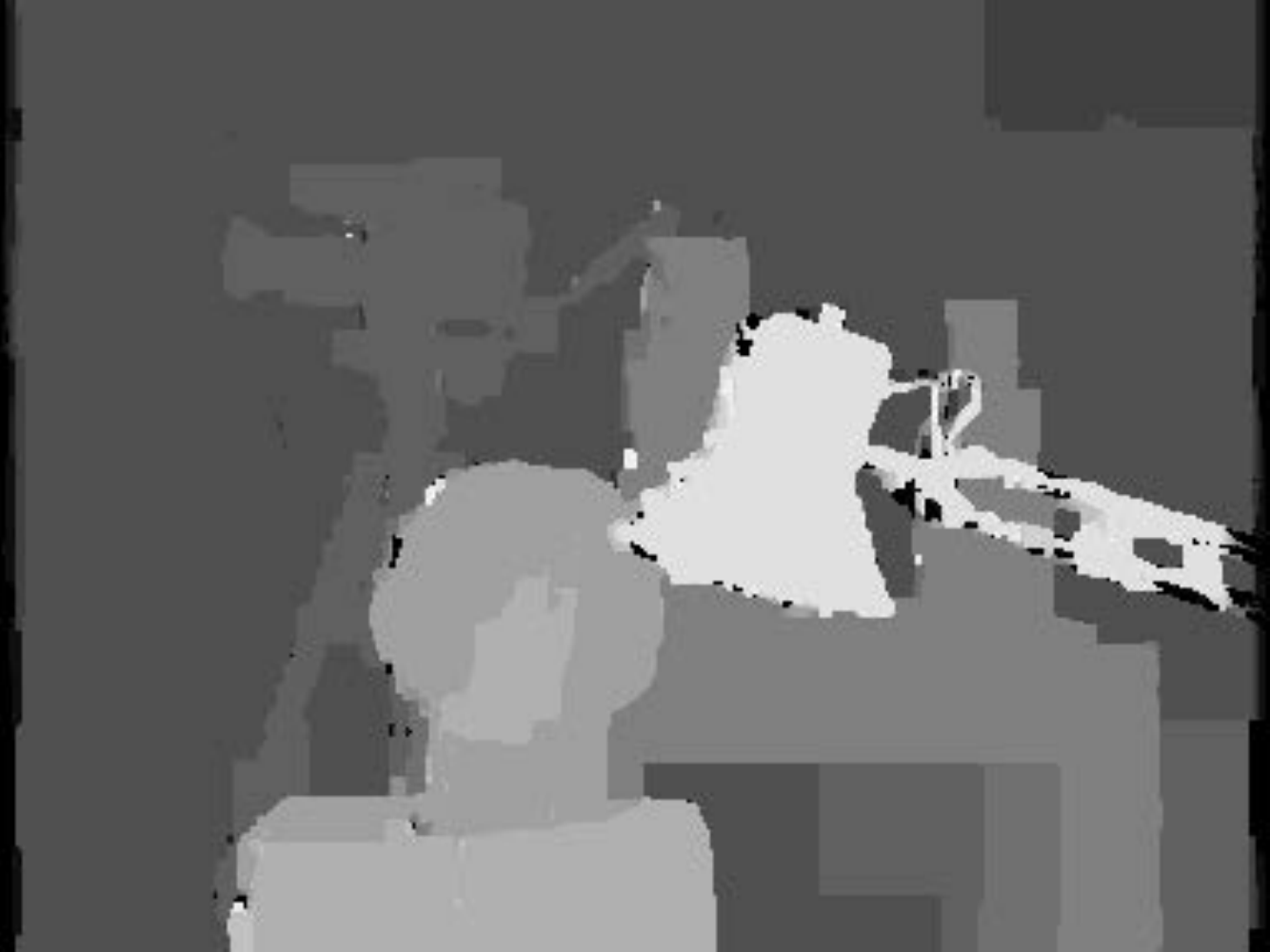}
  \caption{Stereo GC}
\end{subfigure}
\begin{subfigure}{0.45\columnwidth}
  \includegraphics[width=1\columnwidth, trim={0cm 0cm 0cm 0cm}, clip]{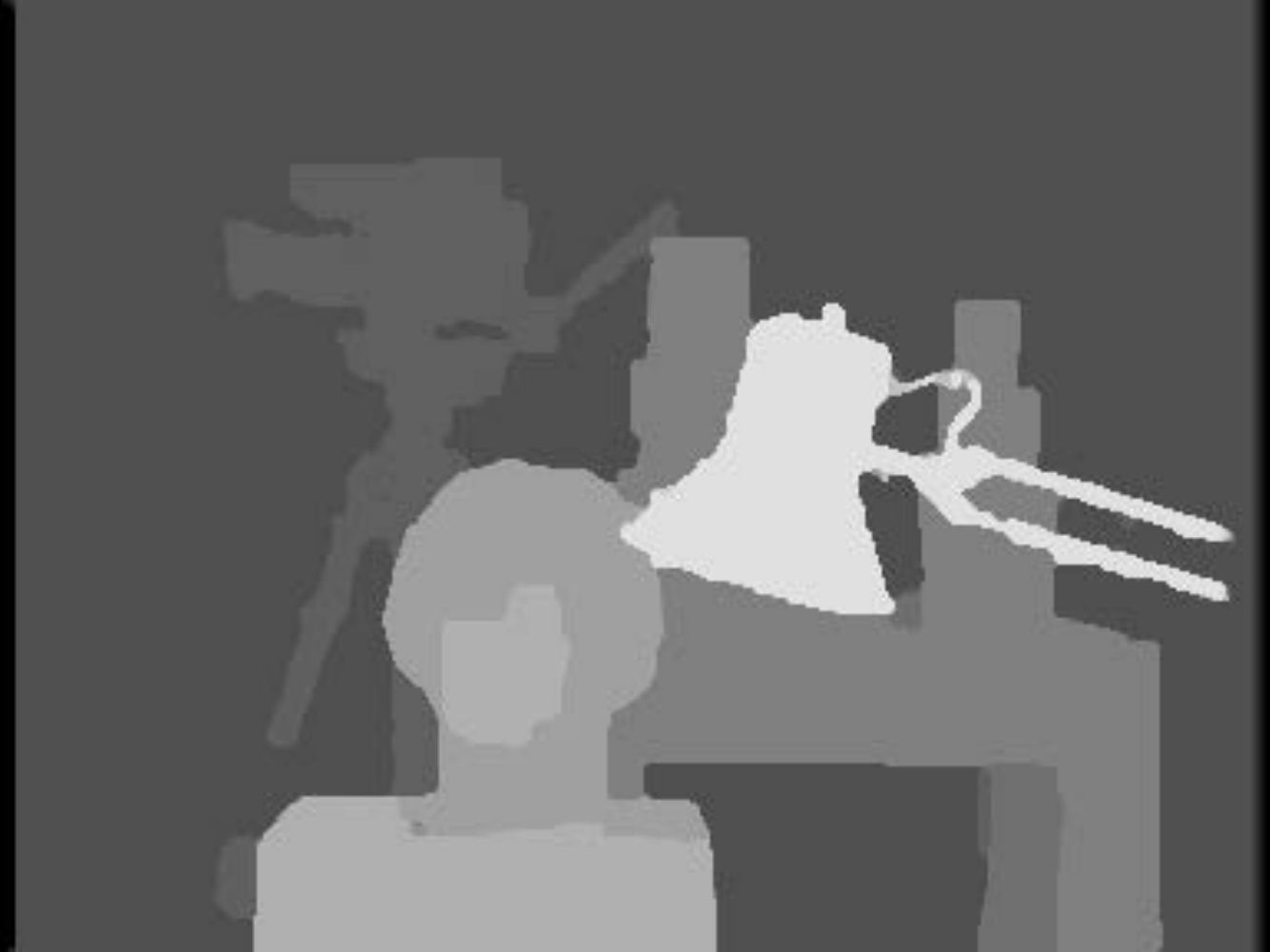}
  \caption{Heuristic Multiscopic GC}
\end{subfigure}
\caption{The disparity maps of Tsukuba obtained by stereo graph cuts and multiscopic graph cuts using heuristic fusion method.}
\label{fig:tsukuba}
% \vspace{-0.1cm}
\end{figure}

% \begin{table}[thb]
% \centering
% \begin{tabular}{c c}
% \toprule
% Method & Bad1 Error \\
% \midrule
% Stereo GC  & 4.06\%\\
% MultiCam GC  & 1.28\% \\
% Geo-consistency MultiCam  & 2.23\%  \\
% Asymmetrical Multiview GC  & 1.30\% \\
% Symmetric Multiview GC & 1.30\% \\
% Our Heuristic Fusion GC & $\mathbf{1.24\%}$\\
% \midrule
% {Stereo MC-CNN} & {$5.48\%$} \\
% {Multiscopic MFuseNet} & {$3.17\%$} \\
% \bottomrule
% \end{tabular}
% \caption{Quantitative results on Tsukuba.}
% \label{tab:tsukuba}
% \vspace{-0.2cm}
% \end{table}

\begin{table}[thb]
\centering
\begin{tabular}{c c}
\toprule
Method & Bad1 Error \\
\midrule
Stereo GC \cite{kolmogorov2014kolmogorov} & 4.06\%\\
MultiCam GC \cite{kolmogorov2002multi} & 1.28\% \\
Geo-consistency MultiCam \cite{drouin2005geo} & 2.23\%  \\
Asymmetrical Multiview GC \cite{wei2005asymmetrical} & 1.30\% \\
Symmetric Multiview GC \cite{maitre2008symmetric} & 1.30\% \\
Our Heuristic Fusion GC & $\mathbf{1.24\%}$\\
\midrule
Stereo MC-CNN & $5.48\%$ \\
Multiscopic MFuseNet & $3.17\%$ \\
\bottomrule
\end{tabular}
\caption{Quantitative results on Tsukuba.}
\label{tab:tsukuba}
\vspace{-0.2cm}
\end{table}

To compare our method with prior multiscopic matching algorithms in a similar setup, we also evaluate our approach on Tsukuba multiscopic data \cite{nakamura1996occlusion}, as is shown in TABLE~\ref{tab:tsukuba}. The disparity value differing from the ground truth by more than one is considered erroneous. This scene has only a small disparity range and sharp edges for thin objects like the lamp, so graph cuts perform better on this image. However, we can still see multiscopic fusion outperforms stereo MC-CNN in this example. The output disparity maps are shown in Fig.~\ref{fig:tsukuba}, in which the multiscopic output is computed using five frames. The result shows that our method achieves the minimum error.

\subsection{Real Robot Experiments}

To show the effectiveness of our approach in real-world robotic applications, we also perform real-world multiscopic experiments. Note that the images captured by our system are not perfectly calibrated and rectified, so there is more noise in the correspondence matching. In this case the multiscopic matching, which is more robust than stereo matching, is in more demand. In our experiments on real robots, we first capture one center image and then capture four surrounding images from the left, right, top, and bottom views. The first example, a toy, is presented in the previous section, and another example is presented in Fig.~\ref{fig:exp_5frame}. 
%
% The fusion of four costs is according to the heuristic rule in equation (\ref{equ:heuristic}). The maximum and minimum searching disparity for these two image sets are the same and set to $70$ and $1$. Because the alignment of these data is not perfect, there is more mismatching and noise. Thus the block size for block matching is set to $17$ and the occlusion penalty $K$ is set to $25$ to encourage the correspondence matching. Other parameters are set as the same as Middlebury datasets.
%
The disparity maps in these two examples clearly show the multiscopic matching reduces a lot of noise in texture-less areas, the occlusion parts, and reflective regions. The reflective metal workpiece, which is everywhere in the industrial environment, can be reconstructed much better.

\section{CONCLUSION}
\label{sec:conclusion}

In this work, we propose a monocular multiscopic vision system for robust depth estimation. A camera mounted at the end of a robot arm is controlled to move in the image plane and take multiple axis-aligned images with the same parallax. We can find pixel correspondences easily because all the captured images are axis-aligned.
We extend stereo matching algorithms to multiscopic algorithms by fusing four cost volumes between the center image and surrounding images, with a new heuristic fusion method and a neural network fusion method. 
The evaluation shows that a more accurate disparity map could be obtained with multiscopic matching compared to stereo matching. 
The noise is significantly reduced on occluded areas and reflective surfaces.
% We hope our work with multiscopic vision can inspire more subsequent works in depth estimation and robotic applications. In the future, we will study different image layouts in the multiscopic vision system.


%%%%%%%%%%%%%%%%%%%%%%%%%%%%%%%%%%%%%%%%%%%%%%%%%%%%%%%%%%%%%%%%%%%%%%%%%%%%%%%%

{\small
\bibliographystyle{IEEEtranN}
\bibliography{ref}

% Generated by IEEEtranN.bst, version: 1.14 (2015/08/26)
\begin{thebibliography}{31}
\providecommand{\natexlab}[1]{#1}
\providecommand{\url}[1]{#1}
\csname url@samestyle\endcsname
\providecommand{\newblock}{\relax}
\providecommand{\bibinfo}[2]{#2}
\providecommand{\BIBentrySTDinterwordspacing}{\spaceskip=0pt\relax}
\providecommand{\BIBentryALTinterwordstretchfactor}{4}
\providecommand{\BIBentryALTinterwordspacing}{\spaceskip=\fontdimen2\font plus
\BIBentryALTinterwordstretchfactor\fontdimen3\font minus
  \fontdimen4\font\relax}
\providecommand{\BIBforeignlanguage}[2]{{%
\expandafter\ifx\csname l@#1\endcsname\relax
\typeout{** WARNING: IEEEtranN.bst: No hyphenation pattern has been}%
\typeout{** loaded for the language `#1'. Using the pattern for}%
\typeout{** the default language instead.}%
\else
\language=\csname l@#1\endcsname
\fi
#2}}
\providecommand{\BIBdecl}{\relax}
\BIBdecl

\bibitem[Biswas and Veloso(2012)]{biswas2012depth}
J.~Biswas and M.~Veloso, ``Depth camera based indoor mobile robot localization
  and navigation,'' in \emph{IEEE International Conference on Robotics and
  Automation (ICRA)}, 2012.

\bibitem[Ye et~al.(2019)Ye, Chen, and Liu]{ye2019tightly}
H.~Ye, Y.~Chen, and M.~Liu, ``Tightly coupled 3d lidar inertial odometry and
  mapping,'' in \emph{IEEE International Conference on Robotics and Automation
  (ICRA)}, 2019.

\bibitem[Yuan et~al.(2019{\natexlab{a}})Yuan, Hang, Song, Kragic, Wang, and
  Stork]{yuan2019reinforcement}
W.~Yuan, K.~Hang, H.~Song, D.~Kragic, M.~Y. Wang, and J.~A. Stork,
  ``Reinforcement learning in topology-based representation for human body
  movement with whole arm manipulation,'' in \emph{IEEE International
  Conference on Robotics and Automation (ICRA)}, 2019.

\bibitem[Yuan et~al.(2019{\natexlab{b}})Yuan, Hang, Kragic, Wang, and
  Stork]{yuan2019end}
W.~Yuan, K.~Hang, D.~Kragic, M.~Y. Wang, and J.~A. Stork, ``End-to-end
  nonprehensile rearrangement with deep reinforcement learning and
  simulation-to-reality transfer,'' \emph{Robotics and Autonomous Systems},
  vol. 119, pp. 119--134, 2019.

\bibitem[Scharstein and Szeliski(2002)]{scharstein2002taxonomy}
D.~Scharstein and R.~Szeliski, ``A taxonomy and evaluation of dense two-frame
  stereo correspondence algorithms,'' \emph{International Journal of Computer
  Vision}, vol.~47, no. 1-3, pp. 7--42, 2002.

\bibitem[Koenderink and Van~Doorn(1991)]{koenderink1991affine}
J.~J. Koenderink and A.~J. Van~Doorn, ``Affine structure from motion,''
  \emph{JOSA A}, vol.~8, no.~2, pp. 377--385, 1991.

\bibitem[Seitz et~al.(2006)Seitz, Curless, Diebel, Scharstein, and
  Szeliski]{seitz2006comparison}
S.~M. Seitz, B.~Curless, J.~Diebel, D.~Scharstein, and R.~Szeliski, ``A
  comparison and evaluation of multi-view stereo reconstruction algorithms,''
  in \emph{IEEE Conference on Computer Vision and Pattern Recognition (CVPR)},
  2006.

\bibitem[Szeliski(2010)]{szeliski2010computer}
R.~Szeliski, \emph{Computer vision: algorithms and applications}.\hskip 1em
  plus 0.5em minus 0.4em\relax Springer Science \& Business Media, 2010.

\bibitem[Kolmogorov and Zabih(2002)]{kolmogorov2002multi}
V.~Kolmogorov and R.~Zabih, ``Multi-camera scene reconstruction via graph
  cuts,'' in \emph{European conference on computer vision}, 2002.

\bibitem[Wei and Quan(2005)]{wei2005asymmetrical}
Y.~Wei and L.~Quan, ``Asymmetrical occlusion handling using graph cut for
  multi-view stereo,'' in \emph{IEEE Conference on Computer Vision and Pattern
  Recognition (CVPR)}, 2005.

\bibitem[Drouin et~al.(2005)Drouin, Trudeau, and Roy]{drouin2005geo}
M.-A. Drouin, M.~Trudeau, and S.~Roy, ``Geo-consistency for wide multi-camera
  stereo,'' in \emph{IEEE Conference on Computer Vision and Pattern Recognition
  (CVPR)}, 2005.

\bibitem[Maitre et~al.(2008)Maitre, Shinagawa, and Do]{maitre2008symmetric}
M.~Maitre, Y.~Shinagawa, and M.~N. Do, ``Symmetric multi-view stereo
  reconstruction from planar camera arrays,'' in \emph{IEEE Conference on
  Computer Vision and Pattern Recognition (CVPR)}, 2008.

\bibitem[Lee and Nguyen(2014)]{lee2014multi}
Z.~Lee and T.~Q. Nguyen, ``Multi-array camera disparity enhancement,''
  \emph{IEEE Transactions on Multimedia}, vol.~16, no.~8, pp. 2168--2177, 2014.

\bibitem[Wilburn et~al.(2005)Wilburn, Joshi, Vaish, Talvala, Antunez, Barth,
  Adams, Horowitz, and Levoy]{wilburn2005high}
B.~Wilburn, N.~Joshi, V.~Vaish, E.-V. Talvala, E.~Antunez, A.~Barth, A.~Adams,
  M.~Horowitz, and M.~Levoy, ``High performance imaging using large camera
  arrays,'' in \emph{ACM Transactions on Graphics (TOG)}.\hskip 1em plus 0.5em
  minus 0.4em\relax ACM, 2005.

\bibitem[Vaish et~al.(2006)Vaish, Levoy, Szeliski, Zitnick, and
  Kang]{vaish2006reconstructing}
V.~Vaish, M.~Levoy, R.~Szeliski, C.~L. Zitnick, and S.~B. Kang,
  ``Reconstructing occluded surfaces using synthetic apertures: Stereo, focus
  and robust measures,'' in \emph{IEEE Conference on Computer Vision and
  Pattern Recognition (CVPR)}, 2006.

\bibitem[Adelson and Wang(1992)]{adelson1992single}
E.~H. Adelson and J.~Y.~A. Wang, ``Single lens stereo with a plenoptic
  camera,'' \emph{IEEE Transactions on Pattern Analysis and Machine
  Intelligence}, no.~2, pp. 99--106, 1992.

\bibitem[Nene and Nayar(1998)]{nene1998stereo}
S.~A. Nene and S.~K. Nayar, ``Stereo with mirrors,'' in \emph{International
  Conference on Computer Vision (ICCV)}, 1998.

\bibitem[Gao and Ahuja(2004)]{gao2004single}
C.~Gao and N.~Ahuja, ``Single camera stereo using planar parallel plate,'' in
  \emph{Proceedings of the 17th International Conference on Pattern
  Recognition, 2004.}, 2004.

\bibitem[Gluckman and Nayar(2002)]{gluckman2002rectified}
J.~Gluckman and S.~K. Nayar, ``Rectified catadioptric stereo sensors,''
  \emph{IEEE Transactions on Pattern Analysis and Machine Intelligence},
  vol.~24, no.~2, pp. 224--236, 2002.

\bibitem[Hu et~al.(2017)Hu, Matsumoto, Takaki, and Ishii]{hu2017monocular}
S.~Hu, Y.~Matsumoto, T.~Takaki, and I.~Ishii, ``Monocular stereo measurement
  using high-speed catadioptric tracking,'' \emph{Sensors}, vol.~17, no.~8, p.
  1839, 2017.

\bibitem[Fehrman and McGough(2014)]{fehrman2014depth}
B.~Fehrman and J.~McGough, ``Depth mapping using a low-cost camera array,'' in
  \emph{Southwest Symposium on Image Analysis and Interpretation}, 2014.

\bibitem[Kolmogorov et~al.(2014)Kolmogorov, Monasse, and
  Tan]{kolmogorov2014kolmogorov}
V.~Kolmogorov, P.~Monasse, and P.~Tan, ``Kolmogorov and zabih's graph cuts
  stereo matching algorithm,'' \emph{Image Processing On Line}, vol.~4, pp.
  220--251, 2014.

\bibitem[Birchfield and Tomasi(1998)]{birchfield1998pixel}
S.~Birchfield and C.~Tomasi, ``A pixel dissimilarity measure that is
  insensitive to image sampling,'' \emph{IEEE Transactions on Pattern Analysis
  and Machine Intelligence}, vol.~20, no.~4, pp. 401--406, 1998.

\bibitem[Ronneberger et~al.(2015)Ronneberger, Fischer, and
  Brox]{ronneberger2015u}
O.~Ronneberger, P.~Fischer, and T.~Brox, ``U-net: Convolutional networks for
  biomedical image segmentation,'' in \emph{International Conference on Medical
  image computing and computer-assisted intervention}, 2015.

\bibitem[Kendall et~al.(2017)Kendall, Martirosyan, Dasgupta, Henry, Kennedy,
  Bachrach, and Bry]{kendall2017end}
A.~Kendall, H.~Martirosyan, S.~Dasgupta, P.~Henry, R.~Kennedy, A.~Bachrach, and
  A.~Bry, ``End-to-end learning of geometry and context for deep stereo
  regression,'' in \emph{Proceedings of the IEEE International Conference on
  Computer Vision}, 2017.

\bibitem[Zbontar and LeCun(2016)]{zbontar2016stereo}
J.~Zbontar and Y.~LeCun, ``Stereo matching by training a convolutional neural
  network to compare image patches,'' \emph{Journal of Machine Learning
  Research}, vol.~17, pp. 1--32, 2016.

\bibitem[Wu et~al.(2018)Wu, Wu, Gkioxari, and Tian]{wu2018building}
Y.~Wu, Y.~Wu, G.~Gkioxari, and Y.~Tian, ``Building generalizable agents with a
  realistic and rich 3d environment,'' \emph{arXiv preprint arXiv:1801.02209},
  2018.

\bibitem[McCormac et~al.(2017)McCormac, Handa, Leutenegger, and
  Davison]{mccormac2017scenenet}
J.~McCormac, A.~Handa, S.~Leutenegger, and A.~J. Davison, ``Scenenet rgb-d: Can
  5m synthetic images beat generic imagenet pre-training on indoor
  segmentation?'' in \emph{Proceedings of the IEEE International Conference on
  Computer Vision}, 2017.

\bibitem[Zhou et~al.(2018)Zhou, Tucker, Flynn, Fyffe, and
  Snavely]{zhou2018stereo}
T.~Zhou, R.~Tucker, J.~Flynn, G.~Fyffe, and N.~Snavely, ``Stereo magnification:
  Learning view synthesis using multiplane images,'' \emph{arXiv preprint
  arXiv:1805.09817}, 2018.

\bibitem[Hirschmuller and Scharstein(2007)]{hirschmuller2007evaluation}
H.~Hirschmuller and D.~Scharstein, ``Evaluation of cost functions for stereo
  matching,'' in \emph{IEEE Conference on Computer Vision and Pattern
  Recognition (CVPR)}, 2007.

\bibitem[Nakamura et~al.(1996)Nakamura, Matsuura, Satoh, and
  Ohta]{nakamura1996occlusion}
Y.~Nakamura, T.~Matsuura, K.~Satoh, and Y.~Ohta, ``Occlusion detectable
  stereo-occlusion patterns in camera matrix,'' in \emph{IEEE Conference on
  Computer Vision and Pattern Recognition (CVPR)}.\hskip 1em plus 0.5em minus
  0.4em\relax IEEE, 1996.

\end{thebibliography}
}
\end{document}